\documentclass[manuscript]{acmart}
\AtBeginDocument{%
  }

\setcopyright{acmlicensed}
\copyrightyear{2026}
\acmYear{2026}
\acmDOI{XXXXXXX.XXXXXXX}
\acmConference[Conference acronym 'XX]{Make sure to enter the correct
  conference title from your rights confirmation email}{June 03--05,
  2026}{Woodstock, NY}
\acmISBN{978-1-4503-XXXX-X/2018/06}

\usepackage{multirow} 
\usepackage{placeins}
\usepackage{caption}

\begin{document}

\title{Simple-to-Complex Structured Demonstrations for Vision-Language-Action Learning}

\author{Xinchuan Qiu}
\authornote{These authors contributed equally to this work.}
\affiliation{%
  \institution{Graduate School of Advanced Science and Engineering, Hiroshima University}
  \city{Higashi-Hiroshima}
  \country{Japan}
}
\email{qiuxinchuan2025@163.com}

\author{Yi Yu}
\authornotemark[1]
\authornote{Corresponding author.}
\affiliation{%
  \institution{Graduate School of Advanced Science and Engineering, Hiroshima University}
  \city{Higashi-Hiroshima}
  \country{Japan}
}

\renewcommand{\shortauthors}{Qiu et al.}

\begin{abstract}

Vision-Language-Action (VLA) models have demonstrated strong capabilities in robotic manipulation by integrating visual perception, language understanding, and robot action generation. Existing research has primarily focused on improving model architectures, training strategies, and dataset scale, while little attention has been paid to how demonstrations are collected and organized. We identify demonstration organization as a fundamental yet overlooked aspect of imitation learning, as it directly affects policy learning efficiency, training stability, and policy generalization. To address this gap, we propose a simple-to-complex structured demonstration collection strategy for VLA learning using a dual-arm robotic platform. Instead of treating demonstrations as independent task trajectories, our approach systematically organizes data through three general principles: (i) decomposing complex manipulation tasks into progressively learnable sub-skills, (ii) standardizing the interaction environment to reduce unnecessary variability, and (iii) organizing demonstrations according to progressively increasing task complexity. This structured design enables VLA models to first acquire fundamental manipulation skills before learning increasingly complex task compositions, facilitating more effective learning of long-horizon manipulation tasks. We evaluate the proposed strategy on two representative robotic manipulation tasks: block grasping and sorting, and towel folding. Experimental results show consistent improvements in task success rate and training stability compared with the baseline method of directly collecting end-to-end complete task trajectories. These findings highlight demonstration organization as a previously underexplored but important factor in VLA learning and provide practical insights into efficient skill acquisition, scalable dataset construction, and long-horizon robotic manipulation.

\end{abstract}




\maketitle

\section{Introduction}

Vision-Language-Action (VLA) models have recently emerged as a promising paradigm for robotic manipulation by jointly modeling visual observations, language instructions, and robot actions. Benefiting from large-scale robotic datasets and foundation models, recent systems such as X-VLA~\cite{XVLA} SmolVLA~\cite{smolvla}, and the $\pi$ series~\cite{pi0,fast,pi05} have demonstrated strong manipulation capabilities and generalization across diverse tasks. 
Despite these advances, existing research has primarily focused on model architecture design, training strategies, and dataset scaling, while comparatively little attention has been paid to how robotic demonstrations are collected and organized. In imitation learning, demonstrations define the supervision signal by inducing the state-action distribution available for policy learning, thereby directly affecting learning efficiency, convergence behavior, and final task performance. As robotic manipulation tasks become increasingly long-horizon and compositional, demonstration organization emerges as a critical yet underexplored dimension of VLA learning.

Through experiments on a dual-arm robotic platform, we observe that different demonstration collection strategies lead to substantially different learning outcomes, even under identical VLA models and training configurations. For block grasping, policies trained on full-task demonstrations often fail to reliably execute intermediate grasping and sorting behaviors. Similarly, in towel folding, learned policies frequently exhibit execution failures or become stuck in intermediate states. These results suggest that naïvely collecting complete task demonstrations may provide suboptimal supervision for long-horizon manipulation learning.

To address this problem, we propose a simple-to-complex structured demonstration collection strategy for VLA learning. Instead of relying solely on full-task trajectories, our approach organizes demonstrations through task decomposition, environment standardization, and progressive complexity scheduling. This structured collection enables VLA models to first acquire fundamental manipulation skills before learning complete task executions, leading to improved learning efficiency and more stable policy optimization.
We evaluate the proposed strategy on two representative dual-arm manipulation tasks, namely block grasping and towel folding. Experimental results show consistent improvements in policy stability and task success rates compared with a direct end-to-end demonstration collection baseline. These findings highlight that demonstration organization is an important design dimension in VLA learning and provide practical guidelines for robotic dataset construction. The main contributions of this work are as follows:

\begin{itemize}

\item We identify demonstration organization as a key yet underexplored factor in VLA learning and demonstrate its importance for learning long-horizon robotic manipulation policies.

\item We propose a simple-to-complex (S2C) structured demonstration collection strategy that organizes demonstrations through task decomposition, environment standardization, and progressive complexity scheduling.

\item We extensively evaluate the proposed strategy on representative dual-arm manipulation tasks involving both rigid and deformable objects, demonstrating consistent improvements in task success rates and learning stability over conventional demonstration collection methods.

\end{itemize}

\section{Related Work}

Existing studies related to this work can be broadly categorized into three directions: (i) demonstration collection for robot learning, (ii) large-scale robotic datasets for VLA learning, and (iii) curriculum learning and task decomposition.

\subsection{Demonstration Collection for Robot Learning}

Robot demonstrations are commonly collected via teleoperation, kinesthetic teaching, or human-guided execution, where full task trajectories are directly recorded as training data~\cite{rt1,bridge,vima,robomimic}. Such approaches are widely adopted in imitation learning~\cite{hussein2017imitation,pomerleau1989alvinn} due to their simplicity and ability to capture complete task executions.

While these full-trajectory collection methods are simple and widely used, prior studies on reinforcement learning from demonstrations~\cite{nair2018overcoming,kalashnikov2018qtopt} have shown that the structure and quality of demonstrations can significantly affect learning efficiency and policy performance. For long-horizon manipulation tasks, directly collecting full trajectories often results in highly coupled and heterogeneous action sequences, which can hinder imitation learning, especially under limited-data regimes. As a result, different demonstration collection strategies may lead to significantly different learning outcomes even when using identical VLA models and training pipelines. Despite these observations, systematic studies on how to structure demonstration collection remain limited~\cite{robomimic}.



In contrast to prior work focusing on model architectures, learning algorithms, or dataset scaling, this paper investigates demonstration organization as a critical design factor in VLA learning. Specifically, we study an S2C structured demonstration collection strategy based on task decomposition, environment standardization, and progressive complexity scheduling, aiming to improve data efficiency and learning stability under limited-data conditions.

\subsection{Large-Scale Robotic Datasets for Vision-Language-Action Learning}

Recent advances in Vision-Language-Action (VLA) learning have been closely associated with the development of large-scale robotic datasets. Early multi-robot datasets such as RoboNet~\cite{robonet} demonstrated the value of aggregating interaction data across different robotic platforms, while Open X-Embodiment~\cite{Open_x_embodiment} further expanded this direction by integrating diverse robot embodiments, tasks, and environments. These datasets provide an important foundation for training generalist robotic policies and have shown that increasing dataset scale and diversity can significantly improve policy robustness and generalization across manipulation tasks.

Built upon such large-scale robotic data, representative VLA systems such as OpenVLA~\cite{openvla} and SmolVLA~\cite{smolvla} further demonstrate the effectiveness of scaling vision-language-action policy learning. These models leverage large and diverse robot demonstration datasets to learn policies that can generalize across a wide range of manipulation scenarios, confirming the central role of large-scale data in developing generalizable robotic manipulation policies.

Recent studies have also investigated how different forms of data scaling and data diversity affect policy learning. The $\pi_{0.5}+\textit{ego}$ framework~\cite{pi05_ego} shows that increasing data diversity enables the model to automatically learn abstract shared features at the level of action intent, thereby enhancing its generalization across different tasks and environments. Similarly, SONIC~\cite{sonic} shows that scaling both data and compute consistently improves policy performance and supports generalization beyond the training distribution.

Despite these successes, collecting large-scale robotic datasets typically requires substantial human effort, multiple robotic platforms, and extensive data collection pipelines. As a result, large-scale dataset construction remains expensive and often impractical for many research settings. In contrast, real-world applications frequently operate under limited-data regimes, where only hundreds or thousands of demonstrations are available, making data efficiency a critical concern.

\subsection{Curriculum Learning and Task Decomposition}

Curriculum learning~\cite{bengio2009curriculum,florensa2017reverse_curriculum} has been widely used to improve optimization by organizing training samples from easy to difficult. In robotics, curriculum-based methods and task decomposition strategies have been extensively explored in reinforcement learning and imitation learning~\cite{vezhnevets2017feudal,d4rl}.

However, existing methods primarily focus on structuring the \emph{training process}, typically assuming that the demonstration dataset has already been collected and fixed. In contrast, how to structure the \emph{data collection process itself} to improve downstream learning remains less explored. This issue is particularly important under limited-data conditions, where dataset quality and organization can be as critical as model design.


\section{Experimental Platform and Task Design}

To evaluate the proposed strategy, we conduct experiments on a real-world dual-arm SO-101 robotic platform. The system consists of two SO-101 manipulators and four synchronized RGB cameras, providing multi-view perception during both data collection and policy execution. All demonstrations are collected via human teleoperation and used for VLA policy fine-tuning. This setup supports a wide range of tabletop manipulation tasks involving both rigid and deformable objects, making it a controlled testbed for studying demonstration organization in VLA learning.

We evaluate our method on two representative long-horizon manipulation tasks: \emph{block grasping and sorting}, and \emph{towel folding}. These tasks are widely used in robotic manipulation research to evaluate compositional skill learning and sequential decision-making under visual and language guidance~\cite{rt1,bridge,vima}. Both tasks require long-horizon, multi-stage manipulation, making them suitable benchmarks for analyzing the effect of demonstration structure.

The block grasping and sorting task requires the robot to locate, grasp, transport, and place objects according to color-based language instructions. The towel folding task involves coordinated bimanual manipulation of deformable objects and sequential execution of folding primitives. The task definitions and their structured demonstration stages are summarized in Table~\ref{tab:ExperimentalTasks}. The stages listed in the table provide a high-level overview of the proposed decomposition, while the detailed definitions and implementation protocols of each stage are presented in Section~\ref{sec:s2c_strategy}.

To ensure a fair comparison, the robotic platform, camera configuration, VLA model architecture, training procedure, and evaluation protocol are kept identical across all experiments. The demonstration collection strategy is the only varying factor, enabling an isolated analysis of its effect on policy learning. Both tasks require the composition of multiple primitive manipulation skills into coherent long-horizon behaviors, which is challenging under limited demonstration data where both skill acquisition and temporal organization are critical~\cite{robomimic}. These properties make them representative testbeds for evaluating structured demonstration collection strategies.

\begin{table*}[t]
\centering
\caption{Long-Horizon Manipulation Tasks and Structured Demonstration Decomposition for Evaluation}
\label{tab:ExperimentalTasks}
\begin{tabular}{p{2.4cm}p{3.5cm}p{3.4cm}p{3.6cm}}
\toprule

\textbf{Task} &
\textbf{Structured Demonstration Stages} &
\textbf{Language Instruction} &
\textbf{Challenges} \\

\midrule

\textbf{Block grasping and sorting} &
Stage 1: Basic Manipulation

Stage 2: Perception-guided manipulation

Stage 3: Dual-arm coordination &
Place each block next to the block of the same color on the table. &
Rigid-object manipulation, language-grounded localization, color-based reasoning, sequential control \\

\addlinespace

\textbf{Towel folding} &
Stage 1: Initial-state handling

Stage 2: State normalization

Stage 3: Rule-based folding &
Please fold the towel neatly on the table. &
Deformable object manipulation, bimanual coordination, long-horizon execution \\
\bottomrule
\end{tabular}
\end{table*}

\section{Simple-to-Complex Structured Demonstration Strategy}
\label{sec:s2c_strategy}

Directly collecting demonstrations from complete long-horizon manipulation tasks often produces complex and highly coupled trajectories that are difficult for VLA models to learn effectively. To address this issue, we propose a simple-to-complex structured demonstration strategy that progressively organizes demonstrations from simple manipulation skills to increasingly complex task executions. 

\subsection{Direct Demonstration Collection}

In the direct demonstration collection setting, each demonstration was recorded as a complete task trajectory, in which the operator executed the entire manipulation process from the initial state to task completion. The resulting end-to-end demonstrations were then directly used for VLA policy learning without further decomposition.

For the block grasping and sorting task, each trajectory covered the complete manipulation workflow, including object localization, object grasping, object transfer, and color-based placement. Similarly, demonstrations for the towel-folding task captured the entire manipulation sequence, including towel grasping, lifting, folding, and final placement.

\begin{figure}[H]
\centering
\setlength{\tabcolsep}{2pt}

\begin{tabular}{cccc}

\multicolumn{4}{c}{\textbf{Block Grasping and Sorting (directly)}} \\
\includegraphics[width=0.23\linewidth]{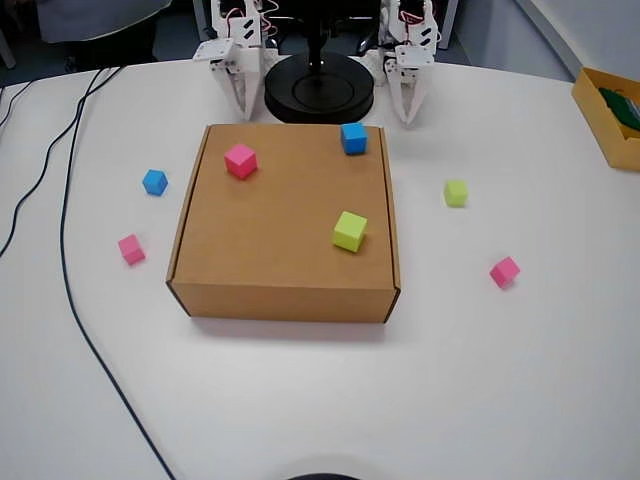}
& \includegraphics[width=0.23\linewidth]{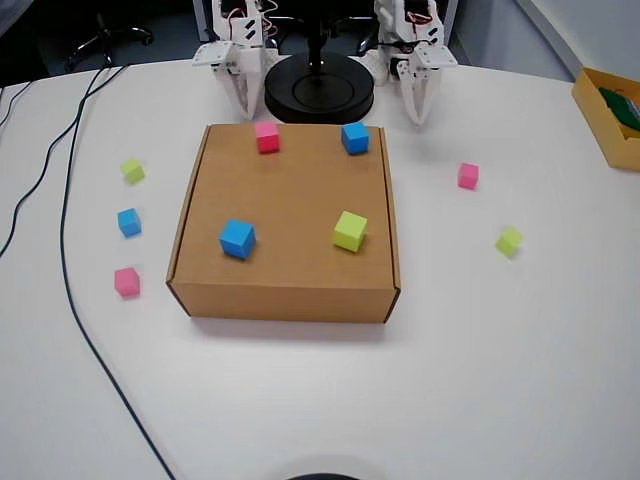}
& \includegraphics[width=0.23\linewidth]{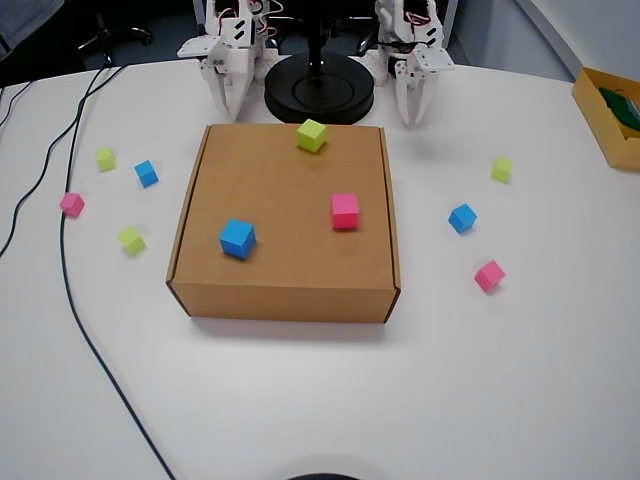}
& \includegraphics[width=0.23\linewidth]{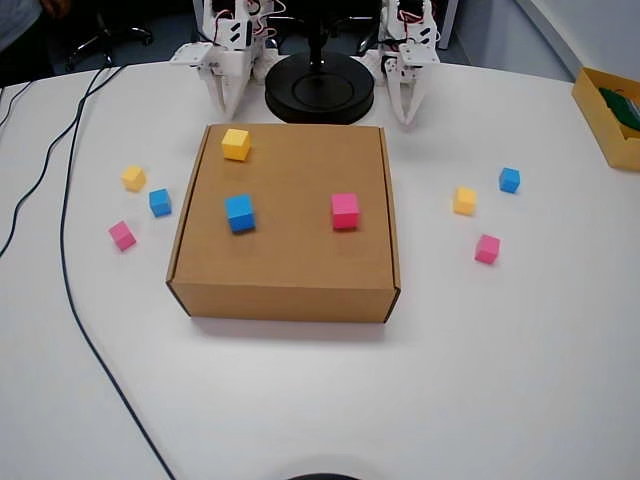}
\\[4pt]

\multicolumn{4}{c}{\textbf{Towel Folding (directly)}} \\
\includegraphics[width=0.23\linewidth]{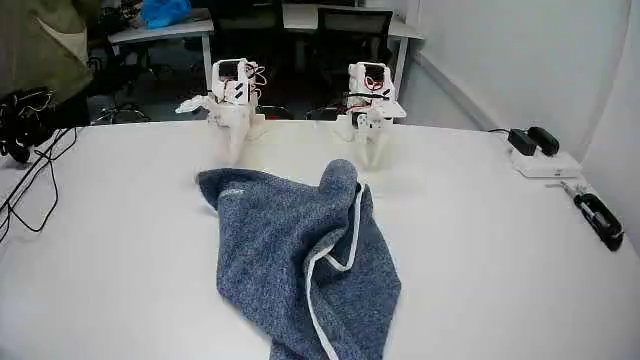}
& \includegraphics[width=0.23\linewidth]{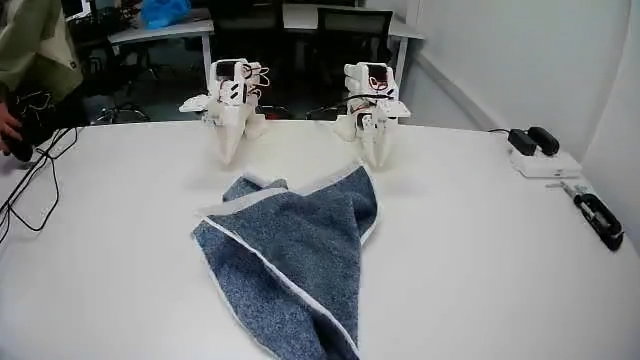}
& \includegraphics[width=0.23\linewidth]{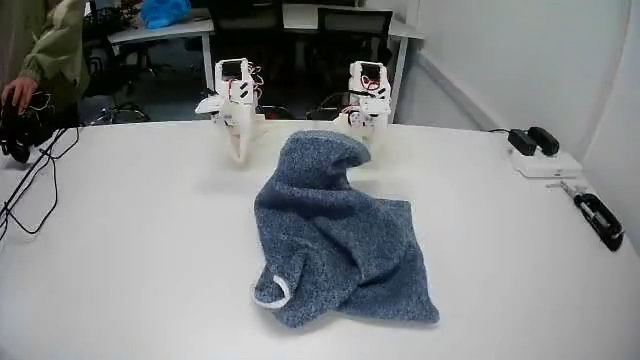}
& \includegraphics[width=0.23\linewidth]{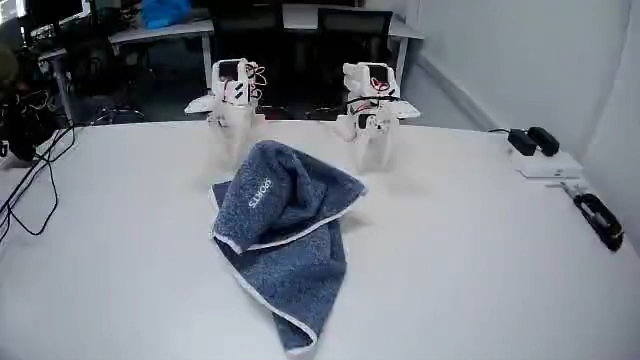}
\\

\end{tabular}

\caption{Examples of directly collected demonstrations for block grasping and sorting, and towel folding. In the direct collection setting, each demonstration is recorded as a single complete trajectory covering the entire task, without being decomposed into simpler sub-skills.}

\label{fig:direct_data}
\end{figure}

This collection protocol follows the conventional paradigm in imitation learning and Vision-Language-Action (VLA) learning, where policies are trained directly from complete task demonstrations without explicitly decomposing complex manipulation tasks into intermediate sub-skills (Figure~\ref{fig:direct_data}). While this end-to-end demonstration strategy is straightforward to implement and has been widely adopted in existing robotic datasets, it often produces highly coupled trajectories for long-horizon manipulation tasks, increasing the difficulty of policy learning under limited-data conditions.

\subsection{Challenges of Direct End-to-End Demonstration Collection}
\label{sec:challenge}

Directly collecting demonstrations from complete long-horizon task executions requires the policy to simultaneously learn primitive manipulation skills and their temporal composition from a limited number of demonstrations. Under limited-data conditions, this coupled learning problem substantially increases the optimization difficulty, as the policy must infer both individual manipulation behaviors and the overall task workflow from the same demonstration trajectories. We consistently observed this limitation across both representative manipulation tasks.

For the block grasping and sorting task, we first collected 200 complete task demonstrations and trained the policy directly on these trajectories. The resulting policy exhibited severe performance asymmetry between the two manipulators. The left arm was able to perform coarse grasping with limited color discrimination, whereas the right arm consistently failed to initiate grasping. Even when a block was manually inserted into the right gripper, the policy was still unable to reliably recognize its color or execute the corresponding placement action. These results suggest that the policy did not acquire balanced manipulation capabilities across the two manipulators and struggled to establish a reliable association between visual color cues and placement actions.

For the towel-folding task, we first collected 200 complete task demonstrations with randomly initialized towel configurations. The learned policy consistently failed to complete the task during evaluation, with the dominant failure mode being prolonged hesitation, where the robot remained stationary without producing meaningful actions. These observations suggest that the policy struggled to infer an effective task progression from the end-to-end demonstrations. Moreover, the large variation in the initial towel configurations introduced substantial visual diversity, making the available demonstrations insufficient to adequately cover the resulting state space.

These observations reveal a common limitation of direct end-to-end demonstration collection. Under limited-data conditions, the policy is required to jointly learn primitive manipulation skills and long-horizon task composition from the same demonstrations. As task complexity increases, the diversity of intermediate states also increases, while the available demonstrations remain insufficient to adequately represent the resulting state space. Consequently, policy optimization becomes considerably more difficult, often leading to unstable execution and poor generalization on complex manipulation tasks.

Motivated by these limitations, we propose a simple-to-complex structured demonstration strategy. Rather than relying solely on complete task demonstrations, our approach progressively organizes demonstrations through task decomposition, environment standardization, and increasing task complexity, enabling VLA models to first acquire prerequisite manipulation skills before learning complete long-horizon manipulation tasks.

\subsection{Overview of the Proposed Strategy}

To overcome the limitations of direct end-to-end demonstration collection, we propose the S2C structured demonstration collection strategy. The core idea of S2C is to avoid learning all required capabilities directly from complete long-horizon task trajectories. Instead, S2C first decomposes a task into several capability objectives with prerequisite relationships, and then designs demonstration collection scenarios around these capability objectives. Specifically, S2C is implemented through three design principles: task decomposition, environment standardization, and progressive complexity scheduling. Task decomposition determines what capabilities the model needs to learn and in what order, while environment standardization and progressive complexity scheduling design the scene conditions for each stage to improve learning stability and generalization.

\textbf{Task decomposition.}
Given a manipulation task $\mathcal{T}$, S2C first decomposes it at the capability level. For many object-centric long-horizon manipulation tasks, task execution can usually be decomposed into three progressive capability objectives:
\[
\mathcal{T}
\Rightarrow
(\mathcal{D}_{\text{manip}}, \mathcal{D}_{\text{state}}, \mathcal{D}_{\text{exec}}),
\]
where $\mathcal{D}_{\text{manip}}$ denotes \emph{basic manipulation}, $\mathcal{D}_{\text{state}}$ denotes \emph{object perception and state understanding}, and $\mathcal{D}_{\text{exec}}$ denotes \emph{task execution}. Basic manipulation focuses on fundamental physical interactions with the target object, such as grasping, lifting, transferring, or placing. Object perception and state understanding focuses on the object attributes, target regions, or intermediate states required for task completion, such as recognizing colors, locating containers, or transforming an object into a standardized state. Task execution combines the first two types of capabilities to complete the full task according to the language instruction.

These three capabilities can form three temporally ordered stages:
\[
\{S_{\text{manip}}, S_{\text{state}}, S_{\text{exec}}\}.
\]
Here, $S_{\text{manip}}$ focuses on learning basic manipulation, $S_{\text{state}}$ introduces object perception and state understanding while preserving basic manipulation, and $S_{\text{exec}}$ further combines the acquired capabilities to complete the full task. Through this decomposition, the policy does not need to simultaneously learn basic actions, task-related perception, state transitions, and long-horizon execution from limited complete-task demonstrations, thereby reducing learning difficulty.

Table~\ref{tab:s2c_general_rule} illustrates how this capability decomposition can be applied to different manipulation tasks. Although the specific objects and task goals differ, their high-level structure is consistent: first learn how to manipulate the target object, then learn task-related object or state information, and finally execute the complete task.

\begin{table}[t]
\centering
\caption{Examples of the general S2C capability decomposition for object-centric manipulation tasks.}
\label{tab:s2c_general_rule}
\begin{tabular}{p{0.30\linewidth} p{0.20\linewidth} p{0.25\linewidth} p{0.17\linewidth}}
\toprule
Task & Basic manipulation & Object perception and state understanding & Task execution \\
\midrule
Put the tissues in the trash can & Grasp tissues & Locate the trash can & Drop tissues into the trash can \\
Put all pens into the pen holder & Grasp pens & Locate the pen holder & Place pens into the pen holder \\
Place each block next to the block of the same color on the table & Grasp blocks & Recognize block colors and color correspondence & Place blocks according to color \\
Fold the towel neatly on the table & Grasp the towel & Normalize the towel to a flat state & Execute the folding sequence \\
\bottomrule
\end{tabular}
\end{table}

\textbf{Environment standardization.}
After task decomposition, S2C further designs the demonstration collection conditions for each stage at the environment level. The goal of environment standardization is to reduce variations that are irrelevant to the current capability within each stage, so that demonstrations can provide more focused supervision for the target capability. For example, when learning basic grasping, variations in object categories, colors, or spatial layouts can be reduced so that the model mainly learns stable grasping. When learning color-conditioned placement, the forms of grasping and placing actions can be kept relatively consistent so that the model mainly learns the correspondence between colors and target locations. When learning towel state normalization, the range of initial-state variations can be controlled so that the model focuses on transforming different configurations into a unified flat state. In this way, demonstrations within each stage become more consistent, reducing the interference of irrelevant environmental variations in policy learning.

\textbf{Progressive complexity scheduling.}
Across different stages, S2C gradually increases task complexity. That is, new sources of difficulty are introduced only after the previous stage has established a relatively reliable capability, such as additional object attributes, more complex target states, dual-arm coordination, deformable-object state variations, or complete task execution. This process allows later stages to build upon stable prerequisite capabilities, rather than requiring the model to learn all components from complex complete trajectories at the beginning. Therefore, progressive complexity scheduling not only reflects the simple-to-complex collection order, but also helps the model transfer acquired capabilities to more complex task scenarios, improving learning stability and generalization.

In this work, we instantiate S2C on two representative long-horizon manipulation tasks: block grasping and sorting, and towel folding. For block grasping and sorting, the policy first learns reliable block grasping and transfer, then learns color perception and color-conditioned placement, and finally executes the complete dual-arm sorting procedure. For towel folding, the policy first learns to grasp the towel, then learns to transform diverse towel configurations into a standardized flat state, and finally executes the folding sequence from this standardized state. These two tasks involve rigid-object and deformable-object manipulation, respectively, but both follow the same S2C organization logic.

\begin{figure}[H]
\centering
\setlength{\tabcolsep}{2pt}
\begin{tabular}{cccc}
\includegraphics[width=0.23\linewidth]{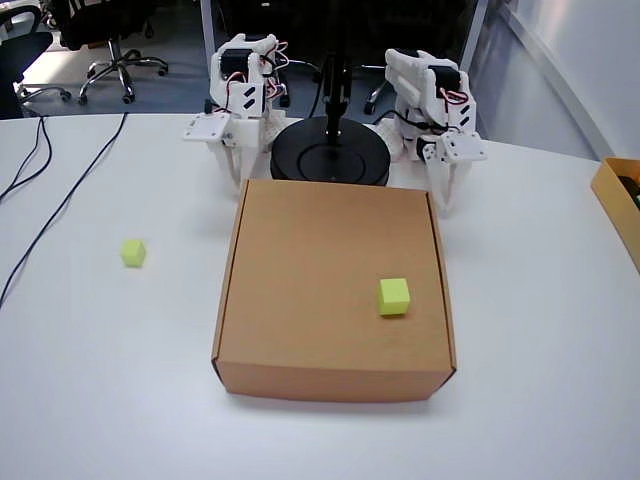}
&
\includegraphics[width=0.23\linewidth]{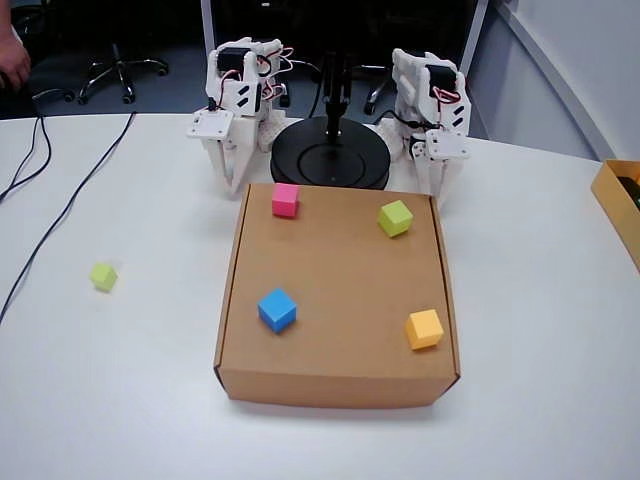}
&
\includegraphics[width=0.23\linewidth]{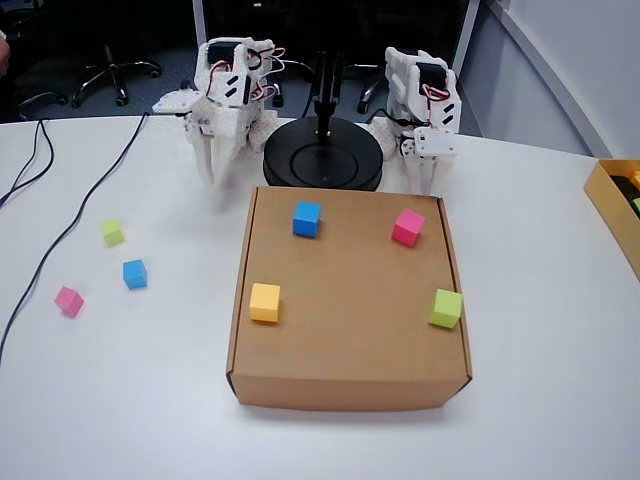}
&
\includegraphics[width=0.23\linewidth]{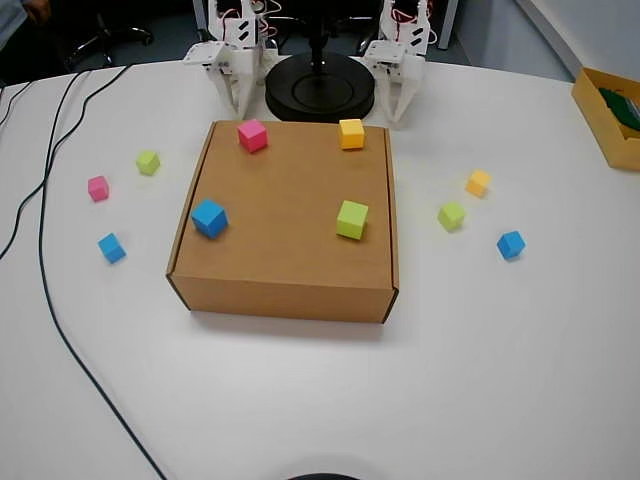}
\\
\end{tabular}
\caption{Representative scene configurations for the block grasping and sorting task under the proposed simple-to-complex strategy. Scene complexity gradually increases from left to right, illustrating the stage-wise demonstration collection process from basic manipulation to object perception and state understanding, and finally to complete task execution.}
\label{fig:simple_to_complex}
\end{figure}

Figure~\ref{fig:simple_to_complex} illustrates representative scene configurations for the block grasping and sorting task under the proposed S2C strategy. From left to right, the manipulation scenario becomes progressively more complex, showing how demonstrations are collected from basic manipulation to object perception and state understanding, and finally to complete task execution. The final stage corresponds to the original end-to-end task, but unlike conventional direct demonstration collection, it is introduced only after the policy has been exposed to simpler prerequisite stages.

The S2C strategy converts demonstration collection from recording independent complete task trajectories into a structured process jointly determined by task capability decomposition and environmental complexity design. Task decomposition reduces the difficulty of learning a complete long-horizon task at once, while environment standardization and progressive complexity scheduling improve policy stability and generalization by controlling and gradually increasing scene variations.

\subsubsection{Block Grasping and Sorting Task}

Following the proposed S2C framework, we instantiate the general capability decomposition for the block grasping and sorting task. Specifically, $\mathcal{D}_{\text{manip}}^{\text{block}}$ corresponds to \emph{basic manipulation}, including reliable block grasping and transfer; $\mathcal{D}_{\text{state}}^{\text{block}}$ corresponds to \emph{object perception and state understanding}, including recognizing block colors and their corresponding placement regions; and $\mathcal{D}_{\text{exec}}^{\text{block}}$ corresponds to \emph{task execution}, including completing the dual-arm color-guided sorting procedure within a shared workspace.

These capability dimensions exhibit a natural dependency hierarchy. Reliable block grasping and transfer form the foundation for color-conditioned placement, while complete dual-arm sorting depends on both stable manipulation and color-based state understanding. Accordingly, S2C organizes demonstration collection into three cumulative stages, denoted by \(S_1 \subset S_2 \subset S_3\), where each stage preserves the capabilities acquired in preceding stages while introducing one additional source of task difficulty.

\textbf{Stage 1 ($S_1$): Basic Manipulation ($\mathcal{D}_{\text{manip}}^{\text{block}}$).}

The first stage focuses on collecting demonstrations to acquire the basic manipulation capability, including reliable block grasping and object transfer. Color-based reasoning and dual-arm task execution are deliberately excluded so that the collected demonstrations capture only fundamental manipulation behaviors. During this stage, blocks of a single color are placed in the workspace, and demonstrations are collected independently for the left and right manipulators. Ten demonstrations are collected for each color on each manipulator across four colors, resulting in a total of 80 demonstrations.

\begin{figure}[!htbp]
\centering
\setlength{\tabcolsep}{2pt}
\begin{tabular}{cccc}
\multicolumn{4}{c}{\textbf{Stage 1 ($S_1$): Basic Manipulation}} \\
\includegraphics[width=0.23\linewidth]{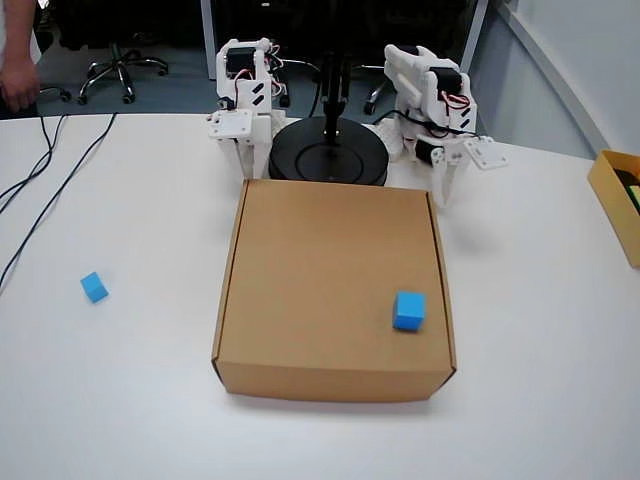}
& \includegraphics[width=0.23\linewidth]{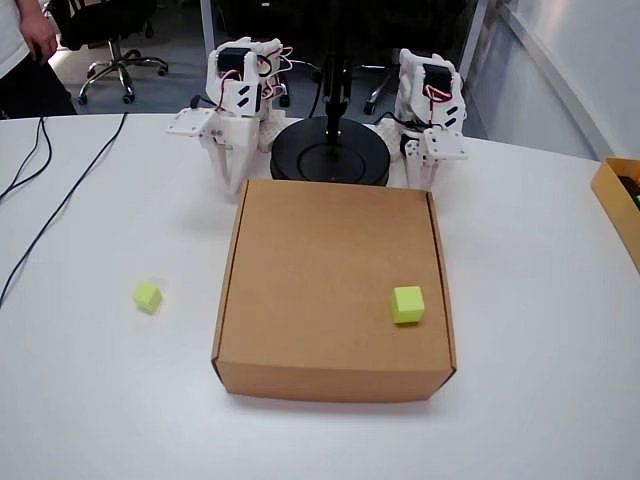}
& \includegraphics[width=0.23\linewidth]{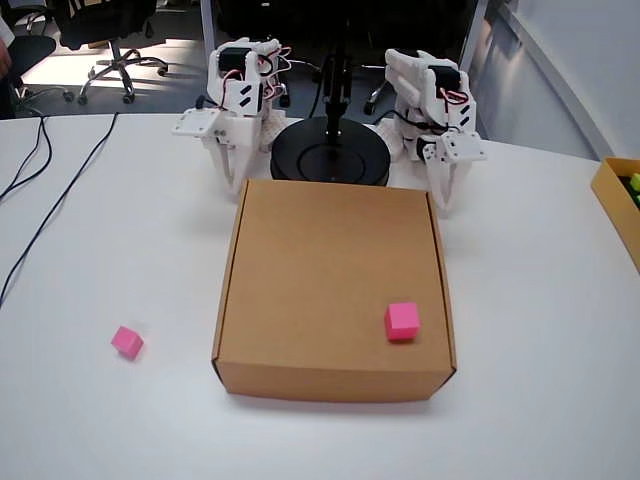}
& \includegraphics[width=0.23\linewidth]{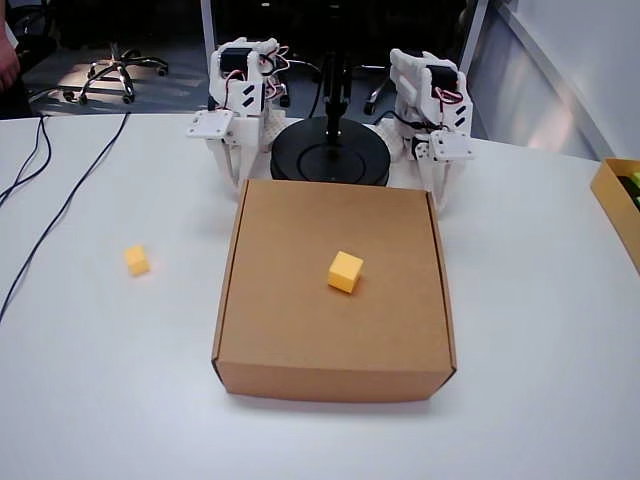}
\\
\end{tabular}
\caption{Basic manipulation under single-color block scenes ($\mathcal{D}_{\text{manip}}^{\text{block}}$).}
\label{fig:block_s1}
\end{figure}

To reduce unnecessary visual variability during the acquisition of basic manipulation skills, the initial scene configuration is standardized in this stage, allowing the demonstrations to focus on robust grasping and placement behaviors. Scene variability and task complexity are then gradually introduced in later stages through additional color combinations, object layouts, and dual-arm execution requirements, improving the adaptability of the policy to more complex scenarios. Representative demonstrations collected in this stage are shown in Figure~\ref{fig:block_s1}.

\textbf{Stage 2 ($S_2$): Object Perception and State Understanding ($\mathcal{D}_{\text{state}}^{\text{block}}$).}

The second stage focuses on collecting demonstrations to acquire the object perception and state understanding capability while preserving the basic manipulation capability established in Stage~$S_1$. During this stage, blocks of multiple colors are placed in the workspace, requiring the robot to associate visual color cues with their corresponding placement regions. Demonstrations continue to be collected independently for the left and right manipulators. Twenty demonstrations are collected for each color on each manipulator across four colors, resulting in a total of 160 demonstrations.

\begin{figure}[!htbp]
\centering
\setlength{\tabcolsep}{2pt}
\begin{tabular}{ccccc}
\multicolumn{5}{c}{\textbf{Stage 2 ($S_2$): Color Perception and Placement}} \\
\includegraphics[width=0.18\linewidth]{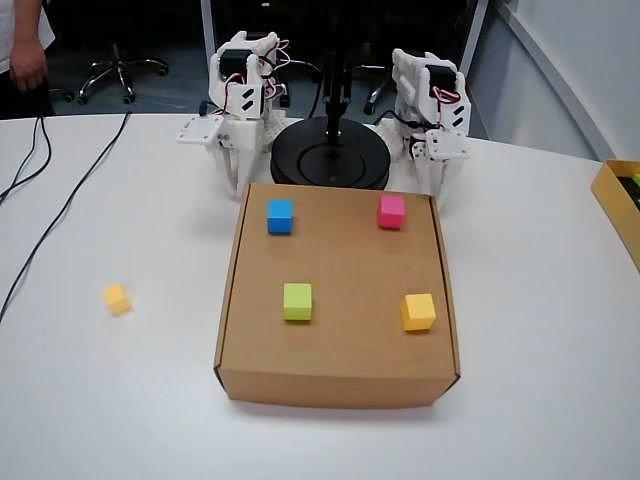}
& \includegraphics[width=0.18\linewidth]{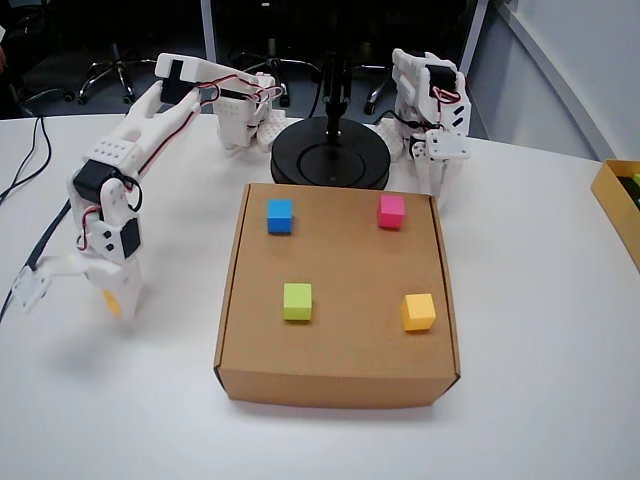}
& \includegraphics[width=0.18\linewidth]{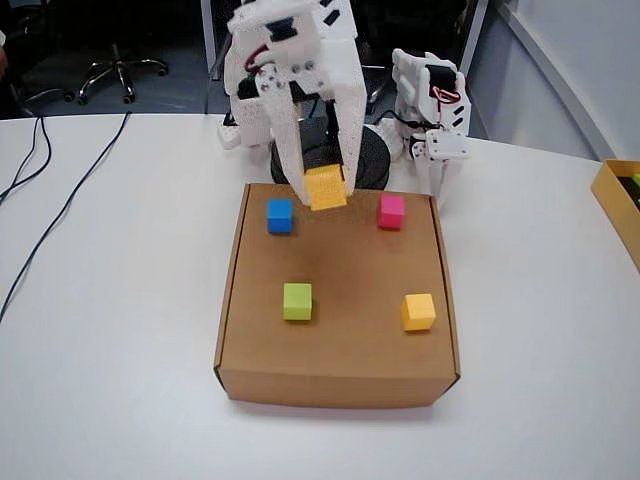}
& \includegraphics[width=0.18\linewidth]{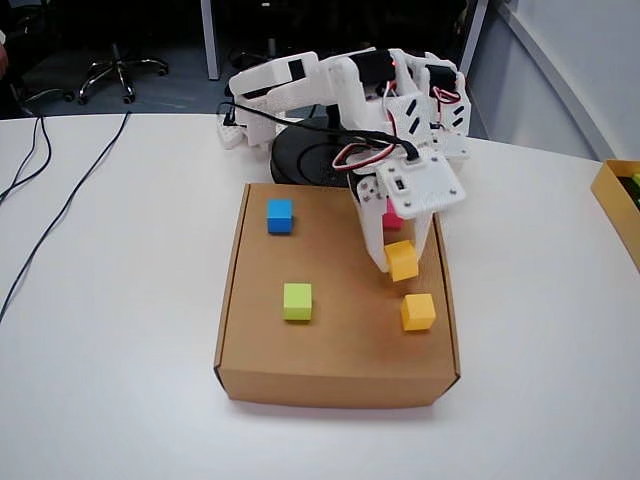}
& \includegraphics[width=0.18\linewidth]{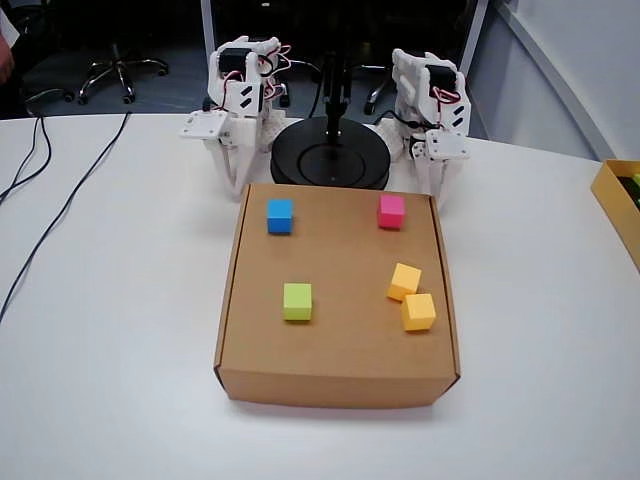}
\\
\end{tabular}
\caption{Color perception and placement under multi-color mixed scenes ($\mathcal{D}_{\text{state}}^{\text{block}}$).}
\label{fig:block_s2}
\end{figure}

Compared with Stage~$S_1$, this stage increases the perceptual complexity of the task while keeping the underlying manipulation primitives unchanged and maintaining standardized experimental conditions. Consequently, the collected demonstrations provide consistent supervision for learning reliable color-conditioned grasping and placement without introducing additional motor execution difficulty. Representative demonstrations collected in this stage are shown in Figure~\ref{fig:block_s2}.

\textbf{Stage 3 ($S_3$): Task Execution ($\mathcal{D}_{\text{exec}}^{\text{block}}$).}

The final stage focuses on collecting demonstrations to acquire the task execution capability while preserving the basic manipulation and object perception capabilities established in Stages~$S_1$ and~$S_2$. During this stage, demonstrations are collected under standardized dual-arm manipulation scenarios, where both manipulators perform block grasping, object transfer, and color-guided placement within a shared workspace. Since the prerequisite manipulation and perception capabilities have already been established, the collected demonstrations focus on completing the full dual-arm sorting procedure, including synchronized motion, spatial coordination, and collision-free cooperation. A total of 60 demonstrations are collected in this stage.

\begin{figure}[!htbp]
\centering
\setlength{\tabcolsep}{2pt}
\begin{tabular}{ccc}
\multicolumn{3}{c}{\textbf{Stage 3 ($S_3$): Dual-Arm Task Execution}} \\
\includegraphics[width=0.30\linewidth]{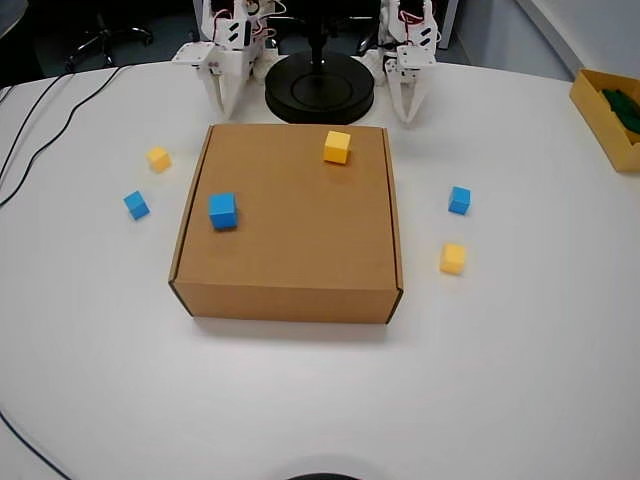}
& \includegraphics[width=0.30\linewidth]{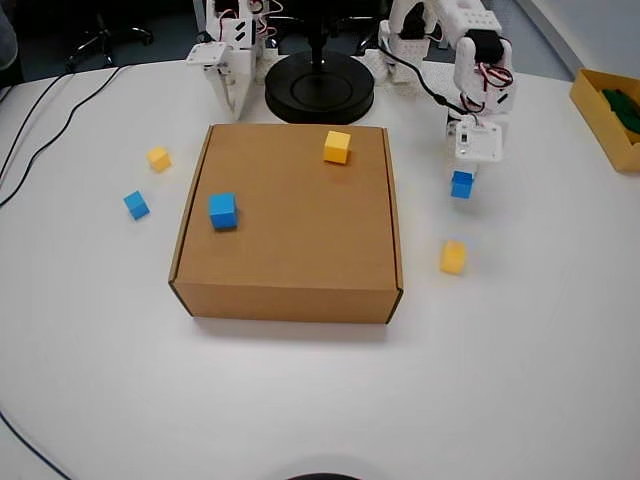}
& \includegraphics[width=0.30\linewidth]{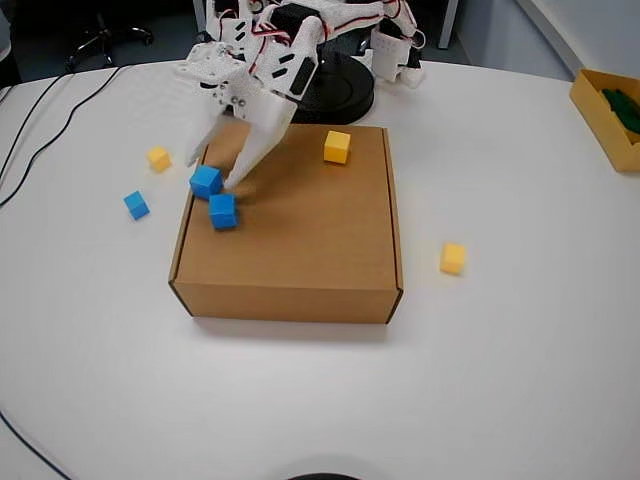}
\\[4pt]
\includegraphics[width=0.30\linewidth]{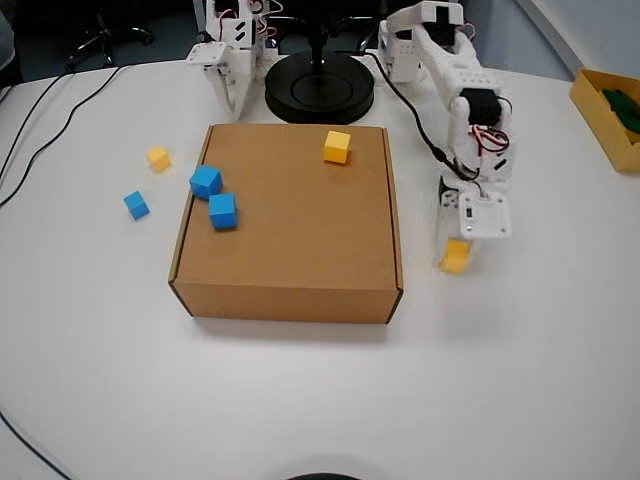}
& \includegraphics[width=0.30\linewidth]{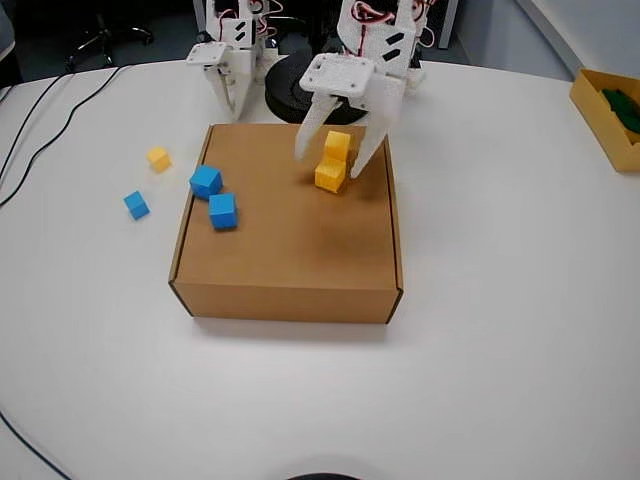}
& \includegraphics[width=0.30\linewidth]{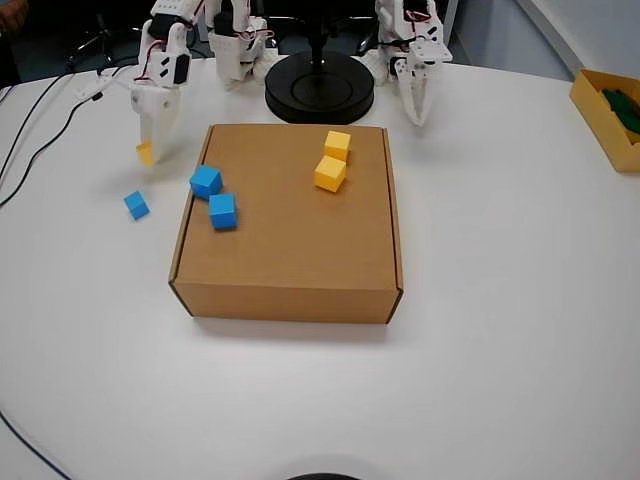}
\\[4pt]
\includegraphics[width=0.30\linewidth]{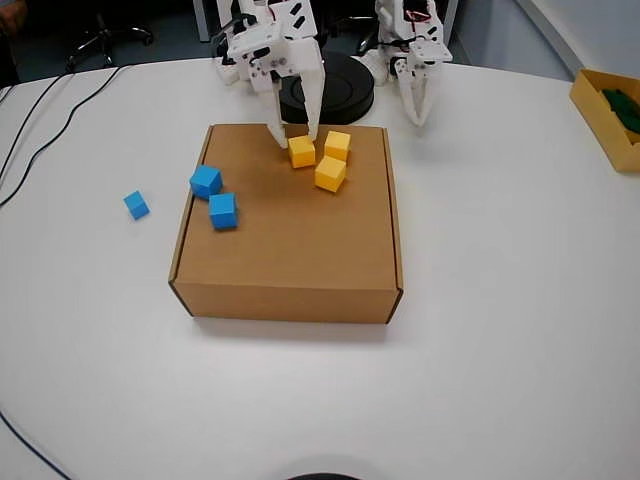}
& \includegraphics[width=0.30\linewidth]{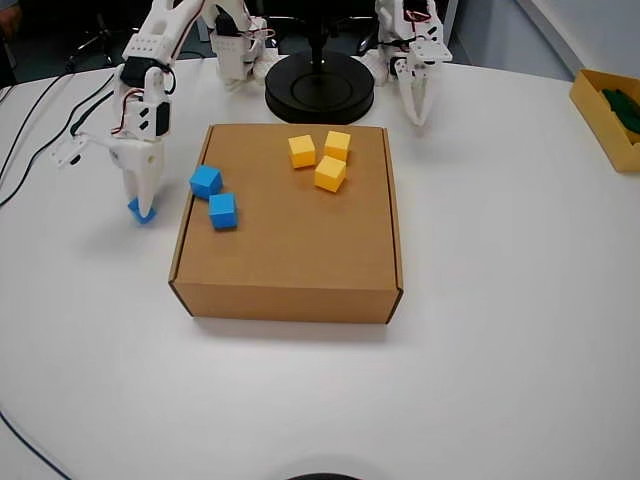}
& \includegraphics[width=0.30\linewidth]{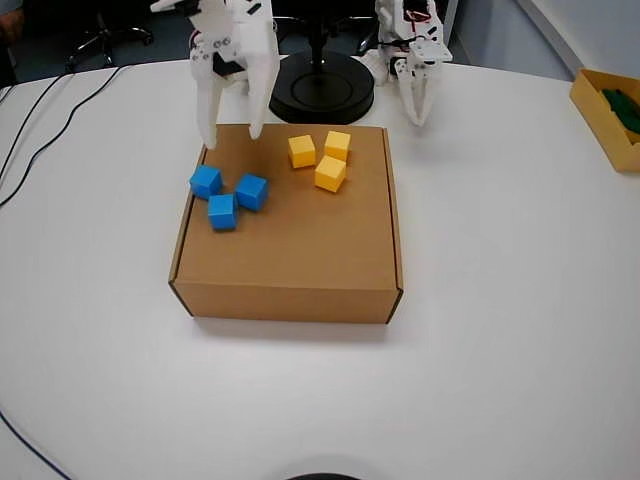}
\\
\end{tabular}
\caption{Dual-arm task execution under cooperative color-guided sorting scenes ($\mathcal{D}_{\text{exec}}^{\text{block}}$).}
\label{fig:block_s3}
\end{figure}

Compared with Stage~$S_2$, this stage introduces complete dual-arm task execution while preserving the same color-guided placement objective and maintaining standardized experimental conditions. Consequently, the collected demonstrations provide consistent supervision for learning reliable task execution without introducing unnecessary perceptual or environmental variability. Representative demonstrations collected in this stage are shown in Figure~\ref{fig:block_s3}.

Overall, the proposed S2C strategy collects a total of 300 demonstrations for the block grasping and sorting task. By progressively introducing basic manipulation, object perception and state understanding, and task execution according to their dependency hierarchy, S2C transforms the original long-horizon manipulation task into a sequence of increasingly challenging demonstration collection stages. Compared with conventional end-to-end demonstration collection, the proposed structured strategy reduces the difficulty of learning multiple interdependent manipulation sub-skills from limited demonstrations and enables more efficient, stable, and robust VLA policy learning.

\subsubsection{Towel Folding Task}

Following the proposed S2C framework, we instantiate the general capability decomposition for the towel-folding task. Specifically, $\mathcal{D}_{\text{manip}}^{\text{towel}}$ corresponds to \emph{basic manipulation}, which is implemented as initial-state handling, including recognizing and grasping the towel from representative initial configurations. $\mathcal{D}_{\text{state}}^{\text{towel}}$ corresponds to \emph{object perception and state understanding}, which is implemented as state normalization, including transforming diverse towel configurations into a standardized flat state. $\mathcal{D}_{\text{exec}}^{\text{towel}}$ corresponds to \emph{task execution}, which is implemented as rule-based folding, including executing the folding sequence from the normalized state.

Unlike rigid-object manipulation, towel folding involves substantial state variability caused by unpredictable deformations of the manipulated object. Consequently, effective policy learning requires the policy to progressively reduce state uncertainty before learning the complete folding procedure. Accordingly, S2C organizes demonstration collection into three progressively more challenging stages, each introducing a single new capability while preserving those acquired in the preceding stages. The corresponding state transition process is formulated as
\begin{equation}
o_0
\xrightarrow{S_1(\mathcal{D}_{\text{manip}}^{\text{towel}})}
o_{\text{grasp}}
\xrightarrow{S_2(\mathcal{D}_{\text{state}}^{\text{towel}})}
o_{\text{flat}}
\xrightarrow{S_3(\mathcal{D}_{\text{exec}}^{\text{towel}})}
o_{\text{final}} .
\end{equation}

\textbf{Stage 1 ($S_1$): Basic Manipulation / Initial-State Handling ($\mathcal{D}_{\text{manip}}^{\text{towel}}$).}

The first stage focuses on collecting demonstrations to acquire the basic manipulation capability for towel folding, namely recognizing and grasping towels from representative initial configurations. To control learning difficulty, demonstrations are initially collected from three typical towel configurations (horizontal, vertical, and slightly disordered), enabling the policy to acquire reliable towel localization and grasping under limited state variability. To further improve generalization, increasingly diverse initial towel configurations are gradually incorporated into the collected demonstrations.

\begin{figure}[!htbp]
\setlength{\abovecaptionskip}{4pt}
\setlength{\belowcaptionskip}{0pt}
\centering
\setlength{\tabcolsep}{2pt}
\begin{tabular}{ccc}
\multicolumn{3}{c}{\textbf{Stage 1 ($S_1$): Basic Manipulation / Initial-State Handling}} \\
\includegraphics[width=0.23\linewidth]{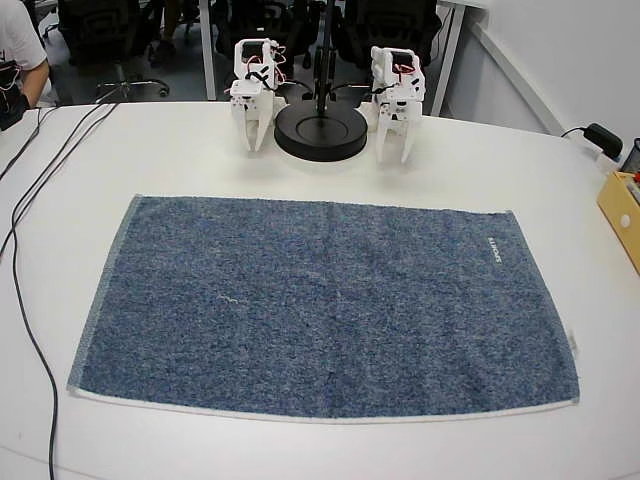}
&
\includegraphics[width=0.23\linewidth]{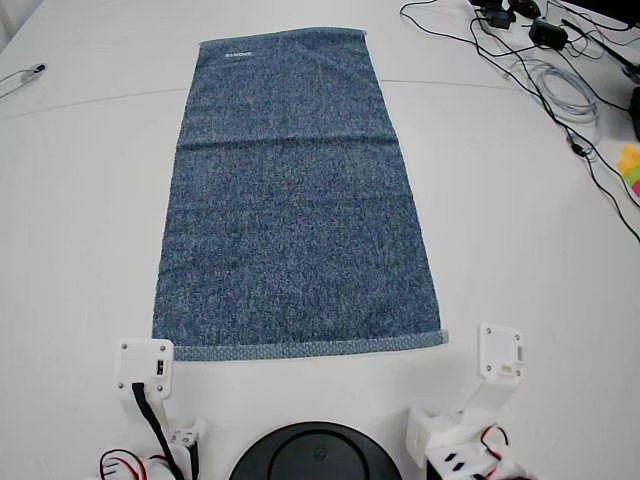}
&
\includegraphics[width=0.23\linewidth]{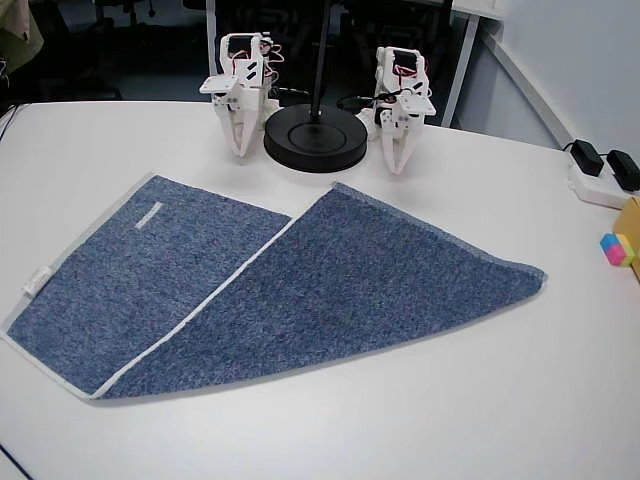}
\\
\end{tabular}
\caption{Representative demonstrations for learning the basic manipulation capability in towel folding, implemented as initial-state handling ($\mathcal{D}_{\text{manip}}^{\text{towel}}$).}
\label{fig:towel_s1}
\end{figure}

Compared with directly collecting demonstrations from arbitrary towel configurations, this stage deliberately limits the diversity of initial states while preserving the same fundamental grasping behavior. Consequently, the collected demonstrations provide consistent supervision for learning reliable towel grasping under standardized conditions. Representative demonstrations collected in this stage are shown in Figure~\ref{fig:towel_s1}.

\textbf{Stage 2 ($S_2$): Object Perception and State Understanding / State Normalization ($\mathcal{D}_{\text{state}}^{\text{towel}}$).}

The second stage focuses on collecting demonstrations to acquire the object perception and state understanding capability while preserving the basic manipulation capability established in Stage~$S_1$. Starting from diverse initial towel configurations, the robot performs a unified unfolding operation that transforms the towel into a standardized flat state, denoted as $o_{\text{flat}}$. Demonstrations are collected under standardized experimental conditions to capture this normalization process.

\begin{figure}[!htbp]
\centering
\setlength{\tabcolsep}{2pt}
\begin{tabular}{cccc}

\multicolumn{4}{c}{\textbf{Stage 2 ($S_2$): State Normalization (Horizontal)}} \\
\includegraphics[width=0.23\linewidth]{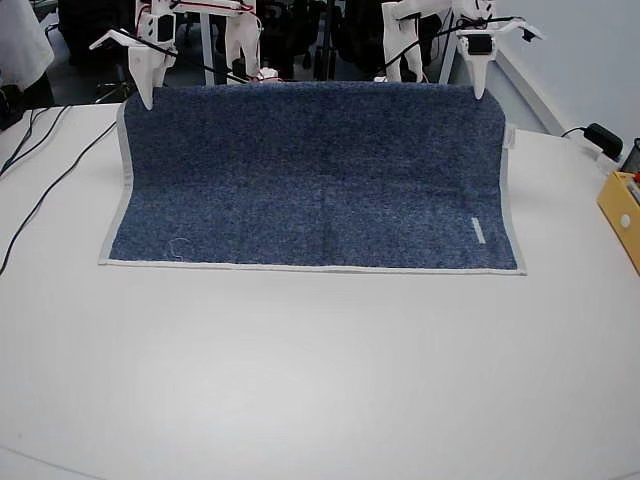}
&
\includegraphics[width=0.23\linewidth]{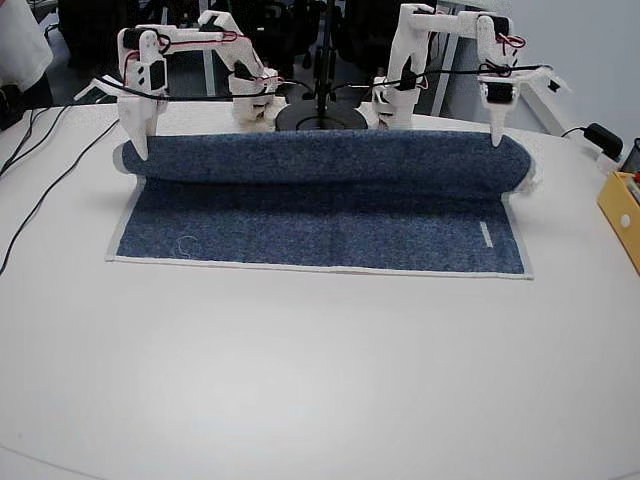}
&
\includegraphics[width=0.23\linewidth]{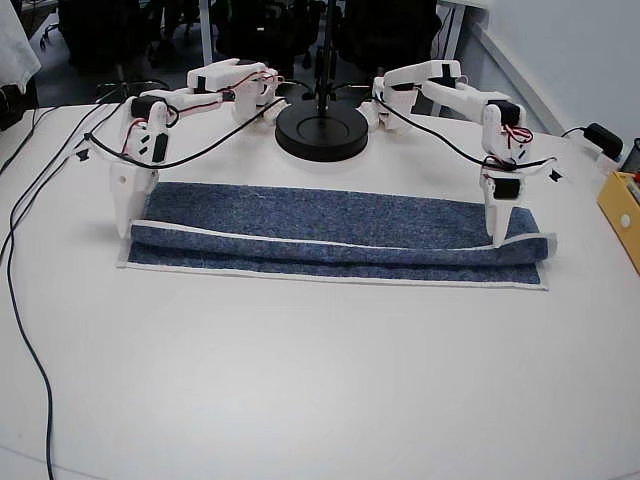}
&
\includegraphics[width=0.23\linewidth]{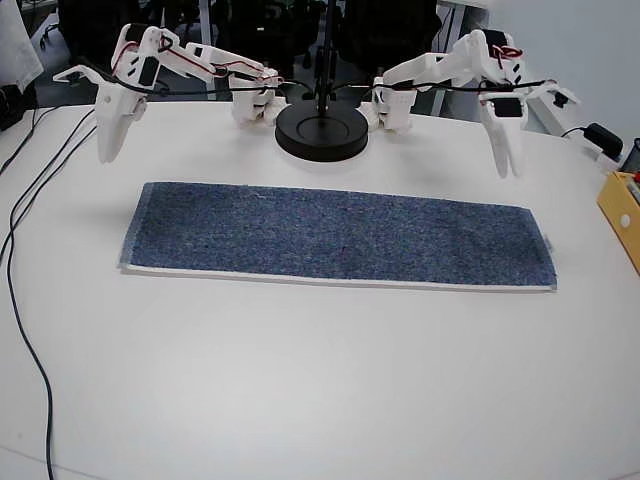}
\\[4pt]

\multicolumn{4}{c}{\textbf{Stage 2 ($S_2$): State Normalization (Vertical)}} \\
\includegraphics[width=0.23\linewidth]{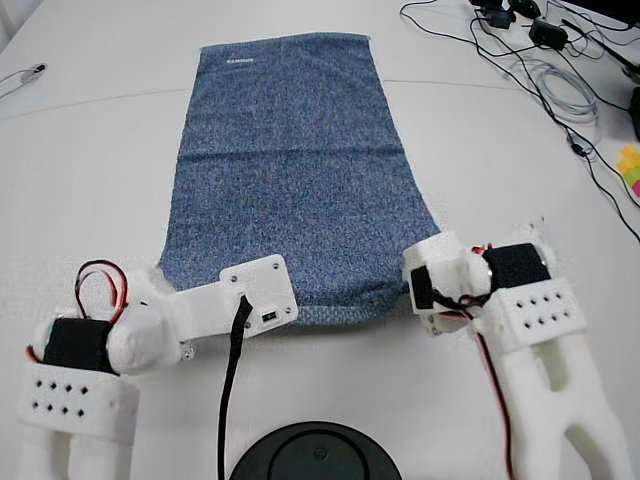}
&
\includegraphics[width=0.23\linewidth]{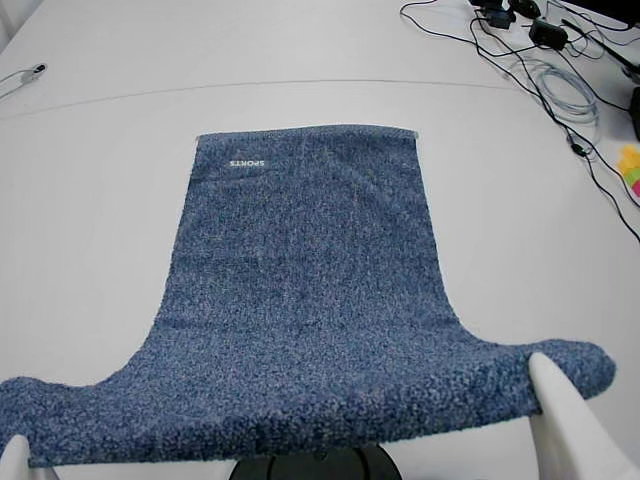}
&
\includegraphics[width=0.23\linewidth]{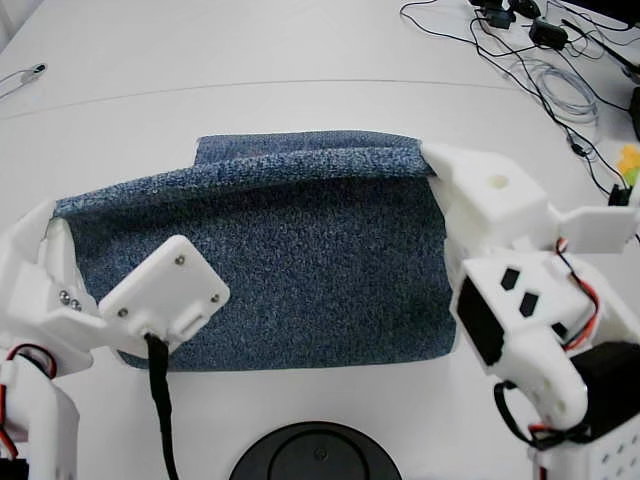}
&
\includegraphics[width=0.23\linewidth]{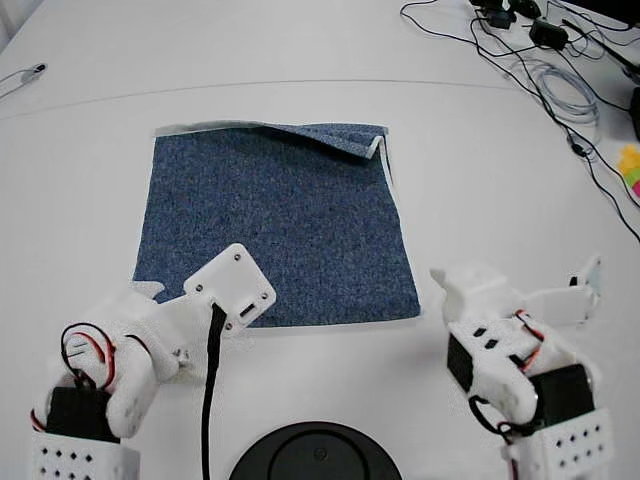}
\\[4pt]

\multicolumn{4}{c}{\textbf{Stage 2 ($S_2$): State Normalization (Slightly Disordered)}} \\
\includegraphics[width=0.23\linewidth]{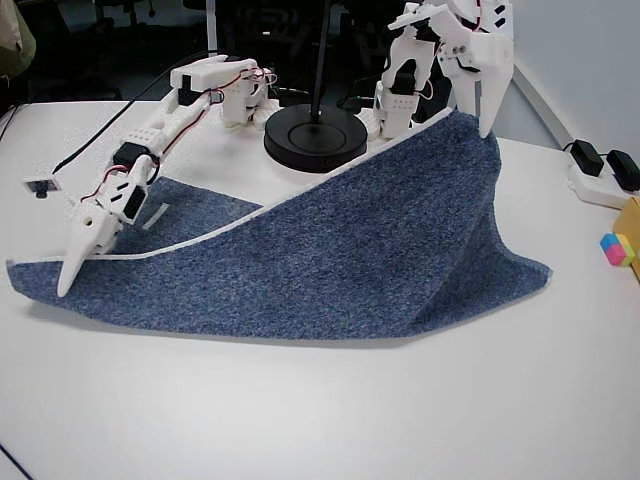}
&
\includegraphics[width=0.23\linewidth]{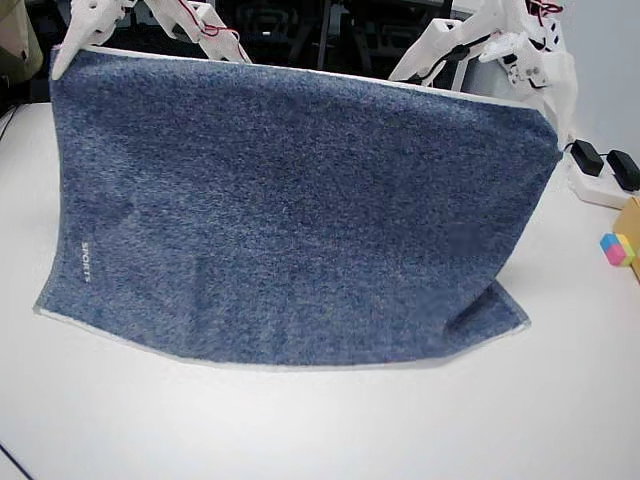}
&
\includegraphics[width=0.23\linewidth]{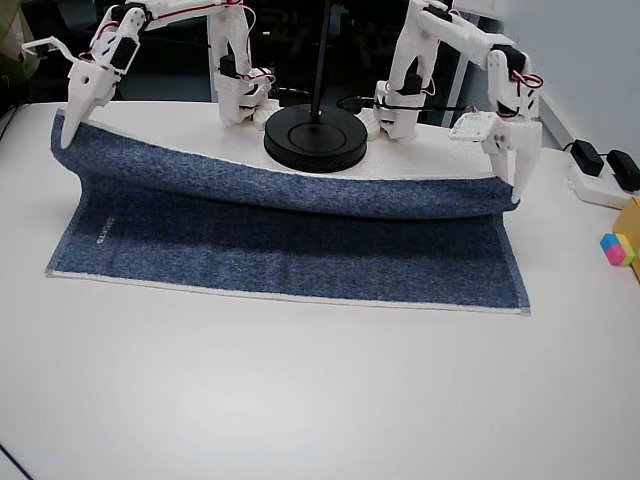}
&
\includegraphics[width=0.23\linewidth]{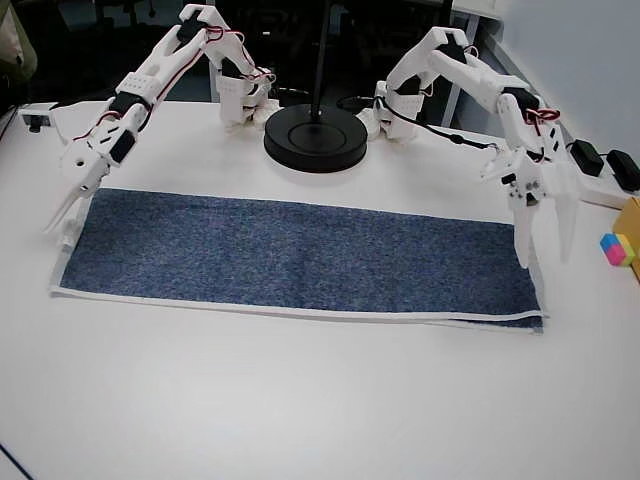}
\\

\end{tabular}
\caption{Representative demonstrations for learning the object perception and state understanding capability in towel folding, implemented as state normalization ($\mathcal{D}_{\text{state}}^{\text{towel}}$), where different initial towel configurations are transformed into the standardized flat state $o_{\text{flat}}$.}
\label{fig:towel_s2}
\end{figure}

Compared with Stage~$S_1$, this stage introduces state understanding and state transformation while preserving the initial towel handling capability. Although the initial observations exhibit substantial variability, the collected demonstrations consistently map diverse towel configurations to the same standardized intermediate state. Consequently, the collected demonstrations provide consistent supervision for learning reliable state normalization while substantially reducing the state variability propagated to the subsequent folding stage. Representative demonstrations collected in this stage are shown in Figure~\ref{fig:towel_s2}.

\textbf{Stage 3 ($S_3$): Task Execution / Rule-Based Folding ($\mathcal{D}_{\text{exec}}^{\text{towel}}$).}

The final stage focuses on collecting demonstrations to acquire the task execution capability while preserving the basic manipulation and state normalization capabilities established in Stages~$S_1$ and~$S_2$. Starting from the standardized flat state $o_{\text{flat}}$, the robot executes a standardized sequence of folding actions to progressively reach the final folded state $o_{\text{final}}$. Since the input state has already been normalized in the preceding stage, the collected demonstrations focus on learning sequential folding behaviors without being directly affected by unpredictable deformable-object configurations.

\begin{figure}[!htbp]
\centering
\setlength{\tabcolsep}{2pt}
\begin{tabular}{ccccc}

\multicolumn{5}{c}{\textbf{Stage 3 ($S_3$): Task Execution / Rule-Based Folding (Horizontal)}} \\
\includegraphics[width=0.18\linewidth]{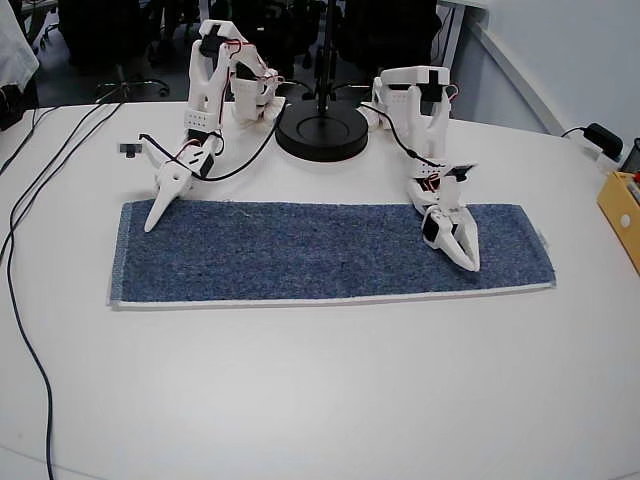}
&
\includegraphics[width=0.18\linewidth]{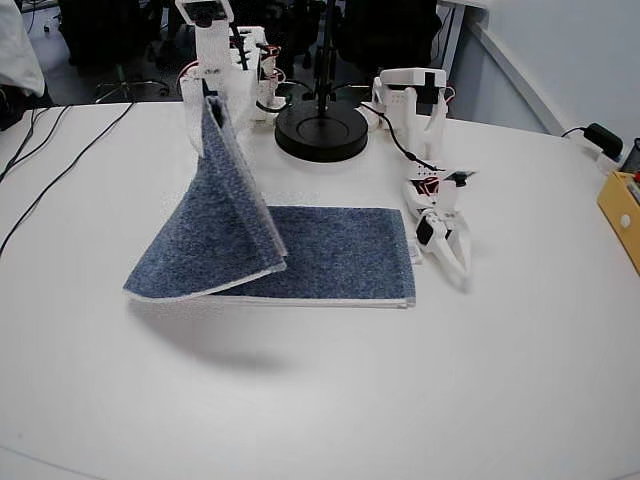}
&
\includegraphics[width=0.18\linewidth]{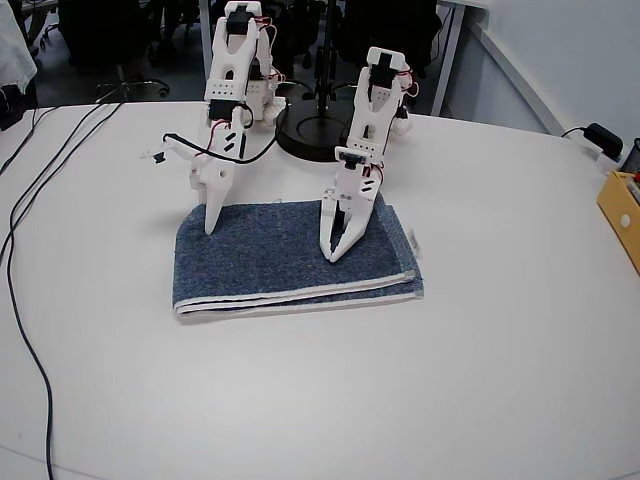}
&
\includegraphics[width=0.18\linewidth]{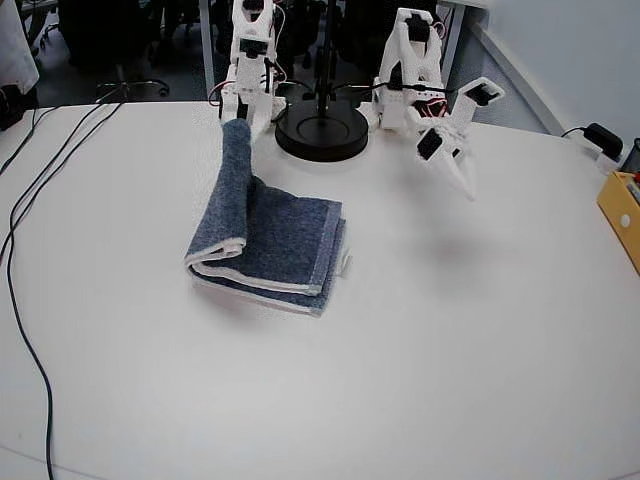}
&
\includegraphics[width=0.18\linewidth]{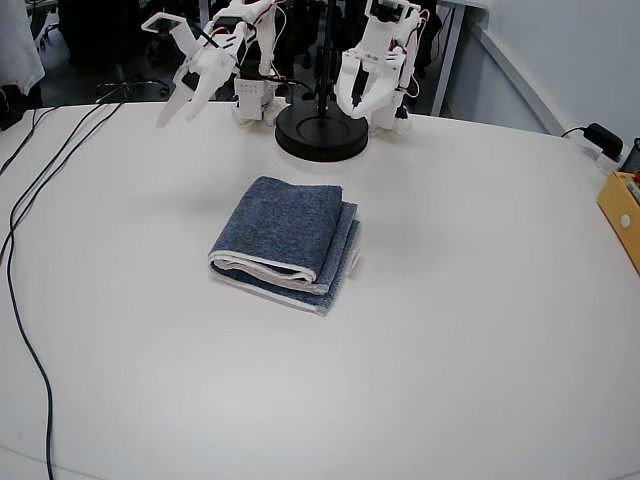}
\\[4pt]

\multicolumn{5}{c}{\textbf{Stage 3 ($S_3$): Task Execution / Rule-Based Folding (Vertical)}} \\
\includegraphics[width=0.18\linewidth]{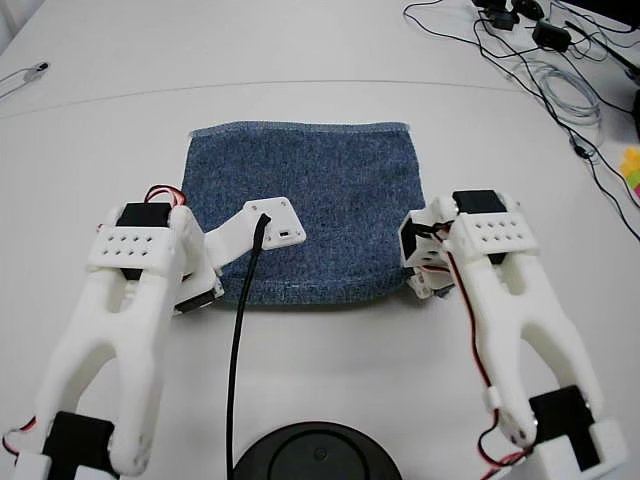}
&
\includegraphics[width=0.18\linewidth]{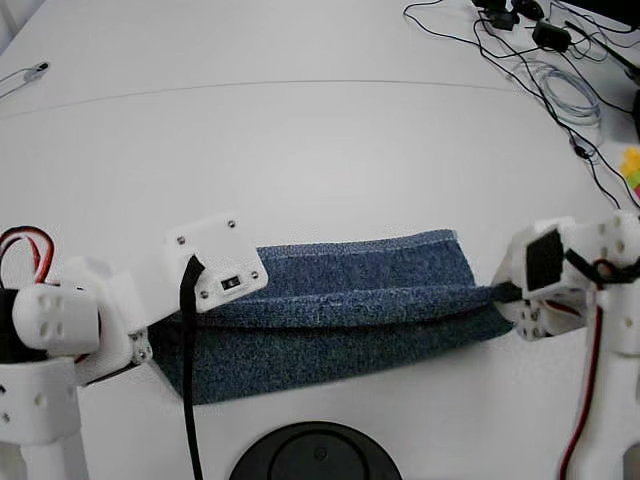}
&
\includegraphics[width=0.18\linewidth]{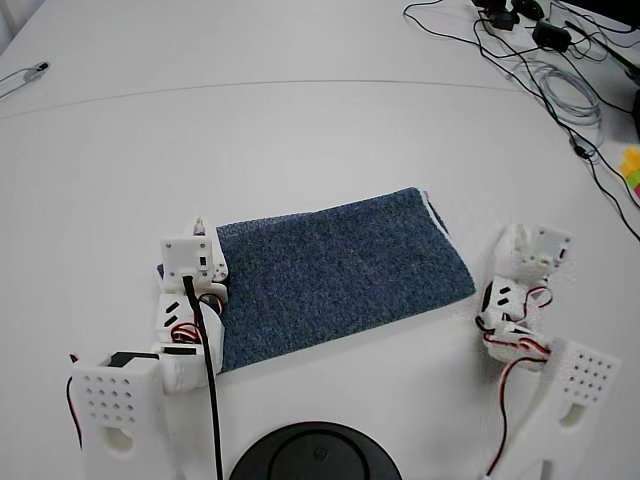}
&
\includegraphics[width=0.18\linewidth]{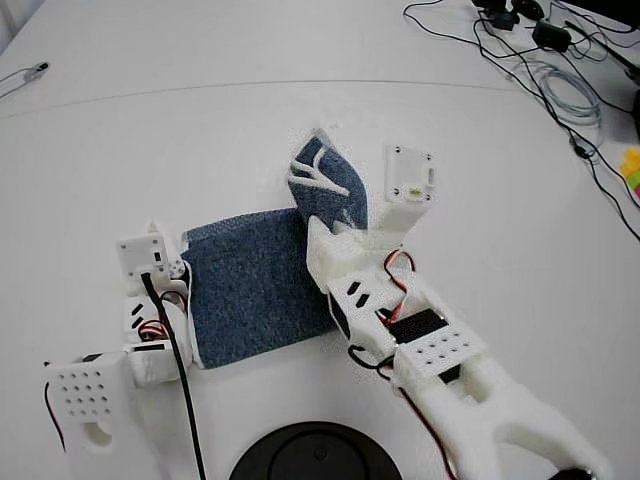}
&
\includegraphics[width=0.18\linewidth]{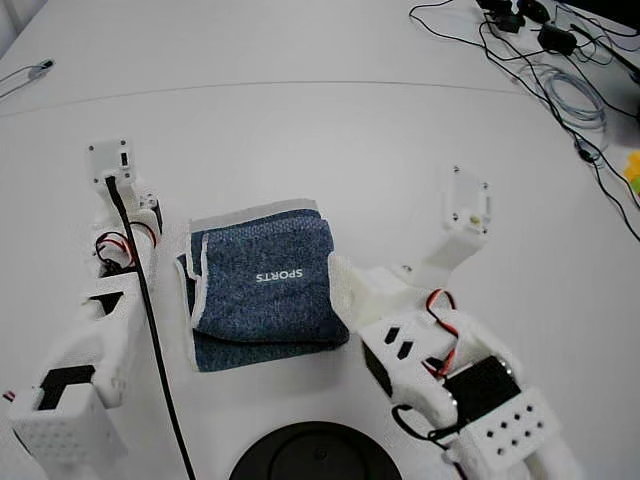}
\\[4pt]

\multicolumn{5}{c}{\textbf{Stage 3 ($S_3$): Task Execution / Rule-Based Folding (Slightly Disordered)}} \\
\includegraphics[width=0.18\linewidth]{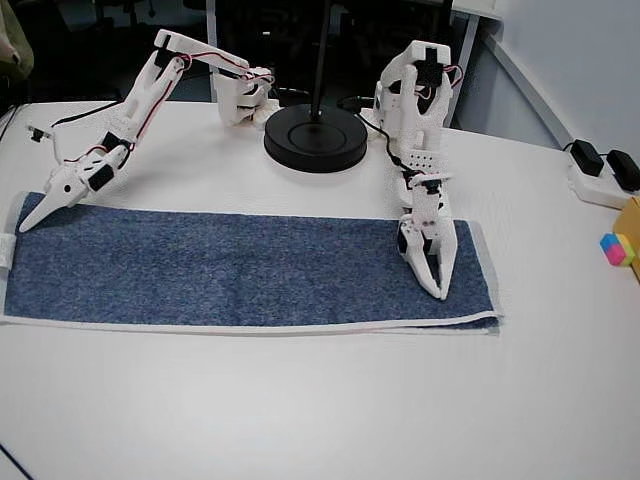}
&
\includegraphics[width=0.18\linewidth]{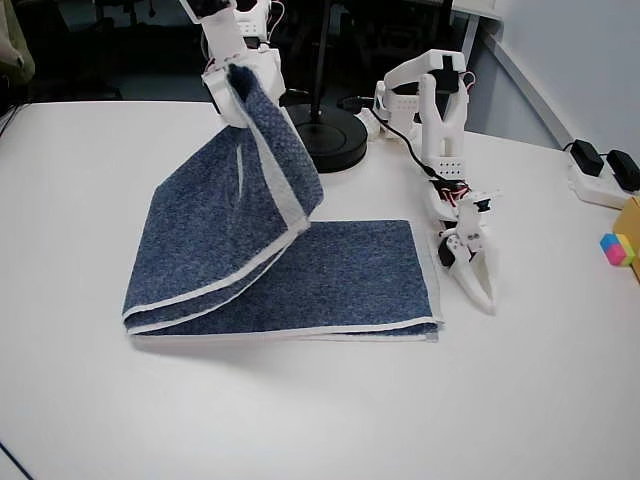}
&
\includegraphics[width=0.18\linewidth]{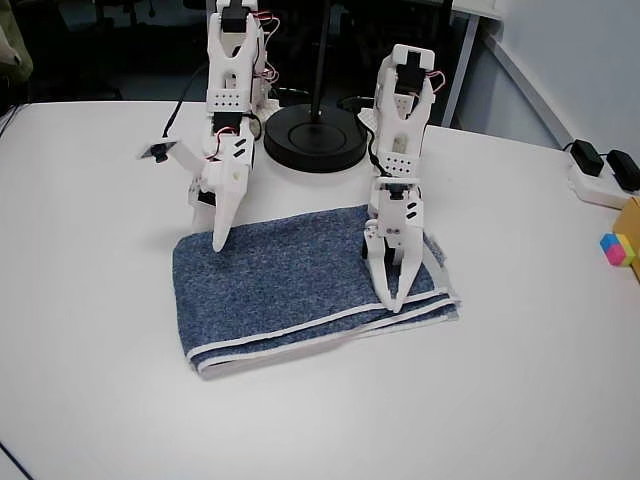}
&
\includegraphics[width=0.18\linewidth]{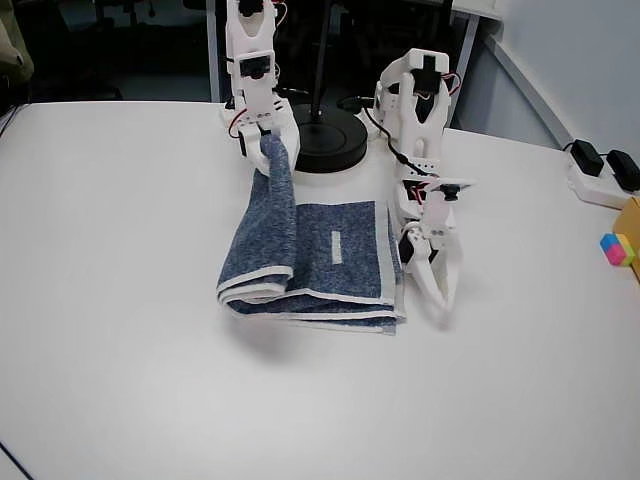}
&
\includegraphics[width=0.18\linewidth]{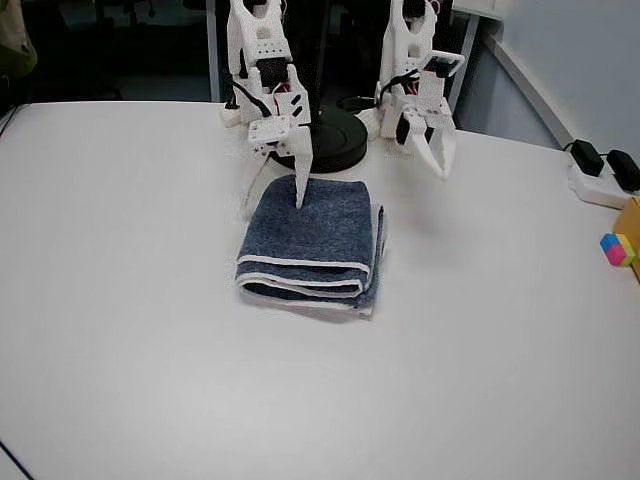}
\\

\end{tabular}
\caption{Representative demonstrations for learning the task execution capability in towel folding, implemented as rule-based folding ($\mathcal{D}_{\text{exec}}^{\text{towel}}$), where the towel is progressively folded from the standardized flat state to the final compact state $o_{\text{final}}$.}
\label{fig:towel_s3}
\end{figure}

Compared with Stage~$S_2$, Stage~$S_3$ shifts the learning focus from state normalization to the complete folding procedure. In this stage, all folding actions start from the standardized flat state $o_{\text{flat}}$, where the uncertainty of the input state has been substantially reduced. As a result, the collected demonstrations provide more consistent supervision for learning reliable long-horizon folding behaviors. Representative demonstrations collected in this stage are shown in Figure~\ref{fig:towel_s3}.

Overall, the proposed S2C strategy organizes demonstration collection for the towel-folding task into three progressively more challenging stages: basic manipulation through initial-state handling, object perception and state understanding through state normalization, and task execution through rule-based folding. By explicitly introducing a dedicated state normalization stage between initial towel handling and final folding, S2C substantially reduces the state variability propagated to subsequent folding actions. Compared with conventional end-to-end demonstration collection, this structured strategy reduces the difficulty of learning multiple interdependent manipulation capabilities under limited demonstration data and enables more efficient, stable, and robust VLA policy learning.

\section{Experimental Results}

In our previous work~\cite{vla_analysis}, we systematically benchmarked several representative VLA models on the low-cost SO-101 robotic platform. Among the evaluated models, $\pi_{0.5}$ consistently demonstrated superior task execution performance, robustness, and generalization across representative manipulation tasks. Based on these findings, we adopt $\pi_{0.5}$ as the underlying policy model throughout this work.

The primary objective of this study is not to compare different VLA architectures, but rather to investigate how structured demonstration collection influences policy learning under identical training conditions. Therefore, all experiments employ the same $\pi_{0.5}$ model, training pipeline, hyperparameter settings, and evaluation protocol. The demonstration collection strategy is the only experimental variable, allowing the effectiveness of the proposed S2C framework to be evaluated independently of differences in model architecture or optimization procedures.

We compare the proposed S2C strategy with conventional direct end-to-end demonstration collection on two representative long-horizon manipulation tasks: block grasping and sorting, and towel folding. For each task, VLA policies are trained using demonstrations collected under the corresponding strategy and evaluated on previously unseen task instances. Experimental results are analyzed from both qualitative and quantitative perspectives, including task execution behavior, policy stability, and task success rates, providing a comprehensive evaluation of the effectiveness of the proposed S2C demonstration collection strategy.

\subsection{Task Success Rate}

Table~\ref{tab:s2c-results} summarizes the task success rates achieved using the two demonstration collection strategies. The task success rate is computed as the ratio between the number of successful trials and the total number of evaluation trials:
\[
\text{Success Rate} = \frac{N_{\text{success}}}{N_{\text{total}}} \times 100\% ,
\]
where \(N_{\text{success}}\) denotes the number of trials in which the task is successfully completed, and \(N_{\text{total}}\) denotes the total number of evaluation trials.

The total number of demonstrations differs between the two methods because the proposed S2C strategy organizes demonstration collection into multiple capability-oriented stages with progressively increasing task complexity. Consequently, the reported results are intended to evaluate the effectiveness of the proposed structured demonstration collection strategy as a whole rather than to compare the two methods using the same number of demonstrations.

\begin{center}
\captionof{table}{Comparison of task success rates under different demonstration collection strategies using $\pi_{0.5}$.}
\label{tab:s2c-results}
\begin{tabular}{llcc}
\toprule
Task & Collection Strategy & \# Demonstrations & Success Rate \\
\midrule

\multirow{2}{*}{Block grasping and sorting}
& Direct & 200 & 0\% (0/5) \\
& Proposed S2C & 300 & 80.0\% (4/5) \\
\midrule

\multirow{2}{*}{Towel folding}
& Direct & 200 & 0\% (0/28) \\
& Proposed S2C & 300 & 25.0\% (7/28) \\
\bottomrule
\end{tabular}
\end{center}

\begin{figure}[H]
\centering
\setlength{\tabcolsep}{2pt}
\begin{tabular}{ccccc}

\multicolumn{5}{c}{\textbf{Block Grasping and Sorting (Success)}} \\
\includegraphics[width=0.18\linewidth]{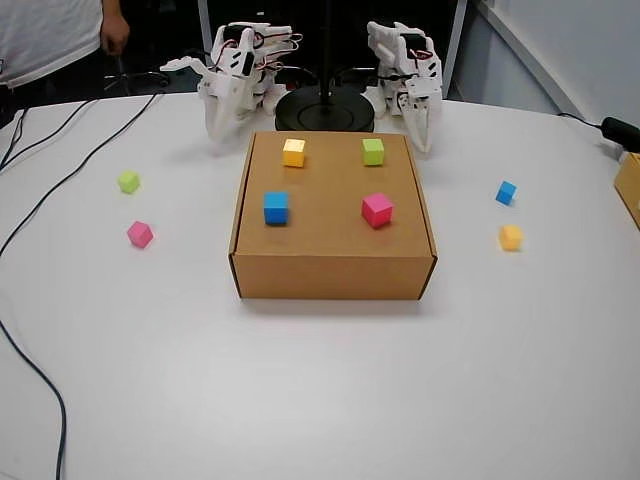}
&
\includegraphics[width=0.18\linewidth]{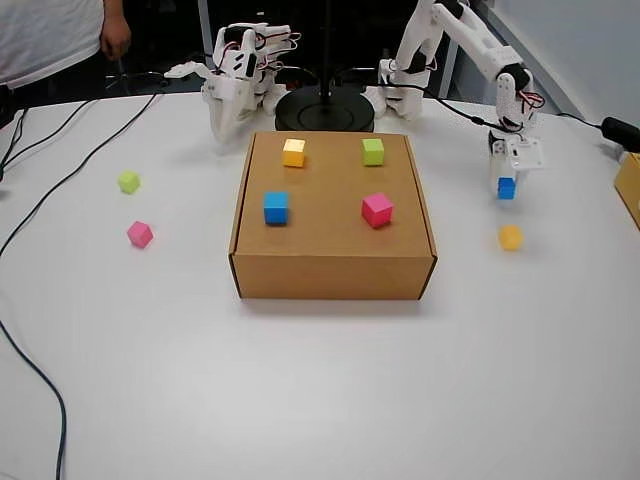}
&
\includegraphics[width=0.18\linewidth]{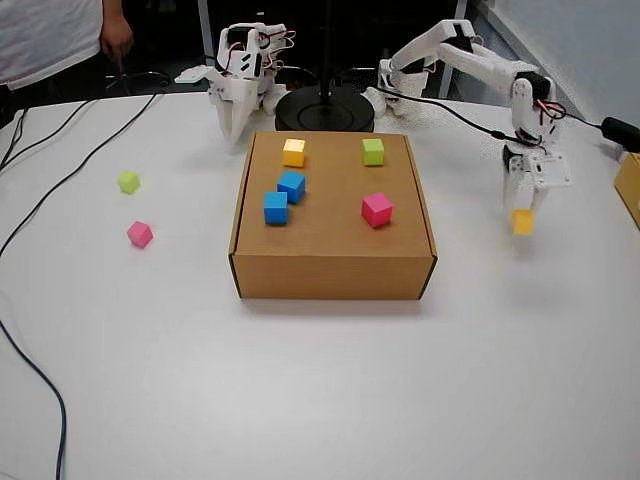}
&
\includegraphics[width=0.18\linewidth]{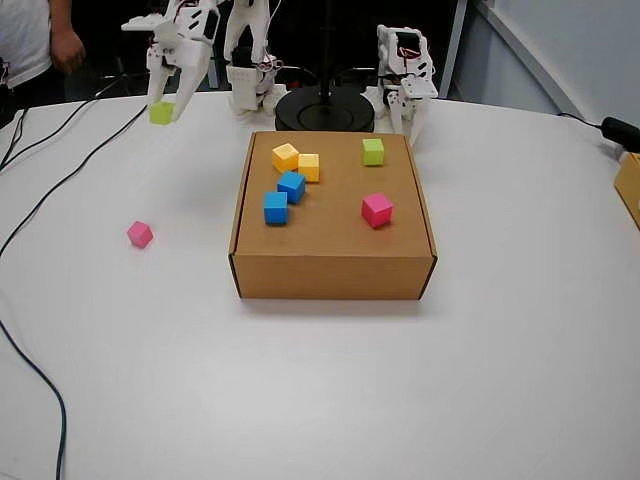}
&
\includegraphics[width=0.18\linewidth]{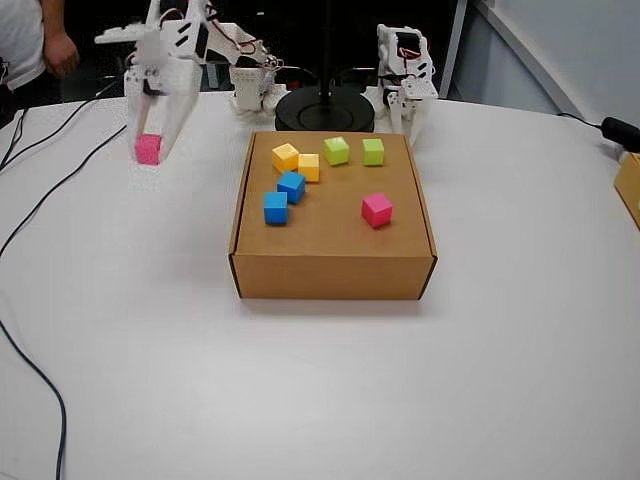}
\\[4pt]

\multicolumn{5}{c}{\textbf{Towel Folding (Success)}} \\
\includegraphics[width=0.18\linewidth]{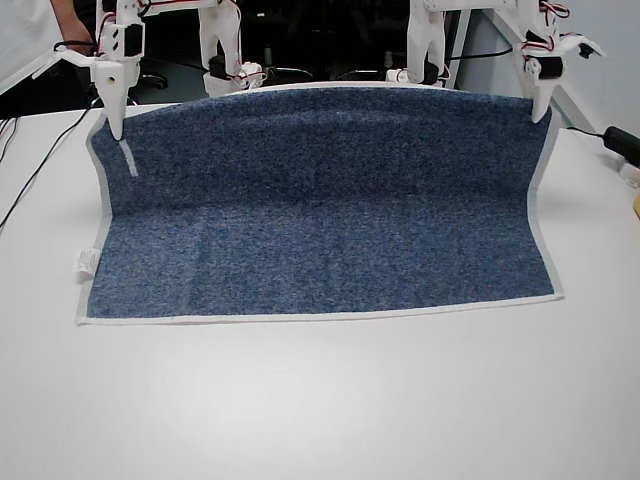}
&
\includegraphics[width=0.18\linewidth]{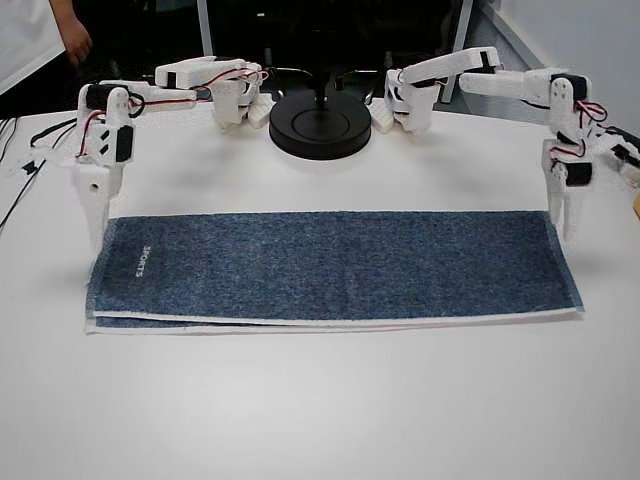}
&
\includegraphics[width=0.18\linewidth]{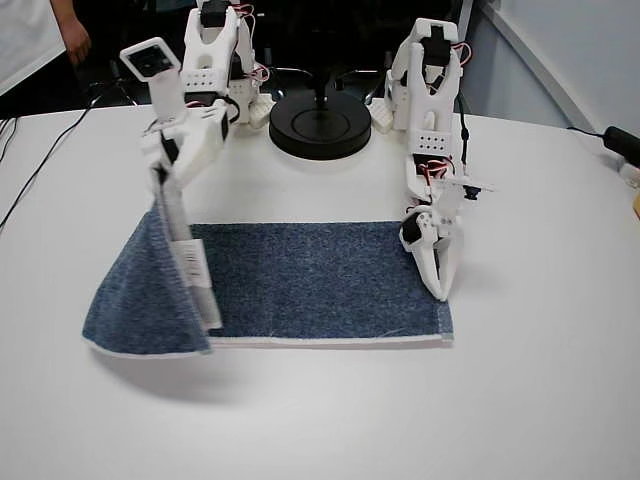}
&
\includegraphics[width=0.18\linewidth]{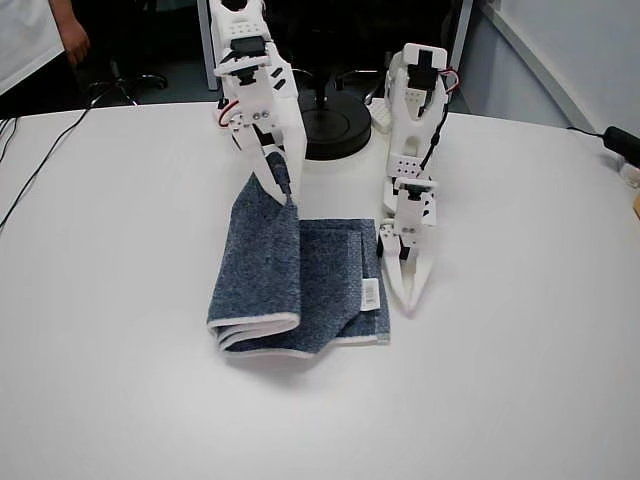}
&
\includegraphics[width=0.18\linewidth]{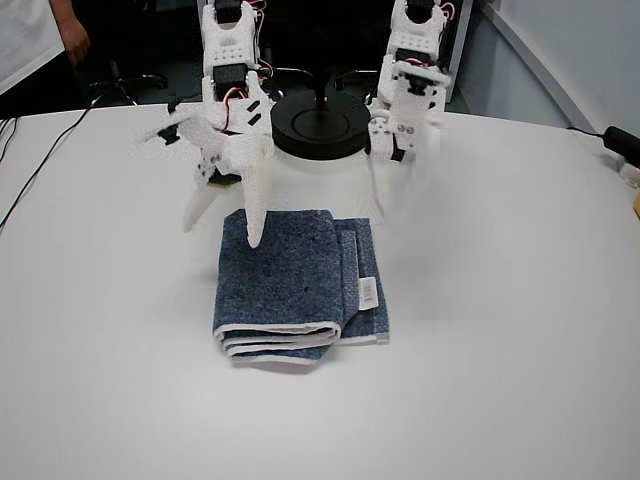}
\\

\end{tabular}
\caption{Representative successful executions after applying the proposed S2C demonstration collection strategy.}
\label{fig:test_result_success}
\end{figure}

In addition to the quantitative results, Figure~\ref{fig:test_result_success} shows representative successful executions obtained using the proposed S2C strategy for both tasks.

For the block grasping and sorting task, the policy trained using directly collected demonstrations failed to complete the task in all evaluation trials, yielding a success rate of 0\%. Consistent with the observations in Section~\ref{sec:challenge}, the learned policy exhibited severe performance asymmetry between the two manipulators and failed to establish a reliable mapping between object colors and their corresponding placement locations. In contrast, the policy trained with the proposed S2C strategy successfully completed four of five evaluation trials, achieving a success rate of 80.0\%. Both manipulators consistently performed object grasping and color-guided placement, indicating that the proposed strategy effectively facilitates the acquisition and integration of multiple interdependent manipulation capabilities.

For the towel-folding task, the policy trained using directly collected demonstrations also failed to complete any evaluation trial, resulting in a success rate of 0\%. The dominant failure mode was prolonged post-grasp hesitation, where the policy failed to proceed with subsequent folding steps after lifting the towel. After adopting the proposed S2C strategy, the policy successfully completed seven of twenty-eight evaluation trials, corresponding to a success rate of 25.0\%. Although the overall success rate remains relatively modest, the policy is able to execute complete folding sequences in multiple trials rather than consistently stalling in the early execution stage. These results suggest that the proposed S2C strategy enables learning of an executable long-horizon manipulation procedure that cannot be reliably obtained through direct end-to-end demonstration collection.

Overall, the proposed S2C demonstration collection strategy consistently outperforms direct end-to-end demonstration collection across both manipulation tasks. Despite the substantial differences between rigid-object and deformable-object manipulation, the proposed strategy improves policy learning in both settings. These findings suggest that, under limited-data conditions, the organization of demonstrations is a critical factor for effective policy learning. By progressively structuring demonstrations according to sub-skill dependencies and task complexity, the proposed strategy enables more effective policy learning and improves long-horizon manipulation performance.

\subsection{Failure Analysis}

Although the proposed S2C demonstration collection strategy improves task success rates, failures still occur during real-world robot evaluation. To further understand the sources of these failures, we analyze the failure cases observed during evaluation. We divide this analysis into two parts: failure definition and failure result analysis.

\subsubsection{Failure Definition}

In this work, a trial is considered successful only if the robot completes the entire task and reaches the final task goal. Otherwise, the trial is counted as a failure. For the block grasping and sorting task, success requires the robot to grasp the target block and place it in the correct target region according to the color instruction. If the robot fails to grasp the block, drops the block during execution, or places the block in an incorrect or unstable region, the trial is regarded as failed. For the towel-folding task, success requires the robot to complete the folding process and produce the expected final folded state. If the robot fails to grasp the towel, loses the towel during manipulation, becomes stuck in an intermediate state, or fails to produce the correct folded configuration, the trial is regarded as failed.

To further characterize the failure cases, we categorize the observed failures into two main types. The first type is \emph{Grasp Failure}, which refers to cases where the robot fails to stably grasp the object. This includes empty grasps, unstable grasps, and cases where the object is dropped or slips out of the gripper during execution. The second type is \emph{Wrong Action}, which refers to cases where the robot performs an action that does not lead to the desired task outcome. For the block task, this includes placing the block outside the correct target region or in an unstable location. For the towel-folding task, this includes incorrect folding actions, incomplete folding, or actions that cause the towel to enter an unintended configuration.

It should be noted that these categories are used to describe the dominant failure events observed during execution. They are not necessarily mutually exclusive; in a single failed trial, multiple failure events may occur sequentially or simultaneously.

For the failure statistics in Table~\ref{tab:s2c_failure_statistics}, the percentage of each failure type is computed as the ratio between the number of events of that failure type and the total number of observed failure events for the corresponding task:
\[
\text{Failure Ratio}_{k} =
\frac{N_{k}}{N_{\text{failure}}} \times 100\% ,
\]
where \(N_{k}\) denotes the number of observed events for failure type \(k\), and \(N_{\text{failure}}\) denotes the total number of observed failure events for the corresponding task. Therefore, the reported percentages indicate the relative proportion of each failure type among all observed failures, rather than the probability of failure over all evaluation trials.

\subsubsection{Failure Result Analysis}

Table~\ref{tab:s2c_failure_statistics} summarizes the main failure events observed under the S2C strategy for the two tasks, providing a quantitative overview of the dominant failure patterns during real-world execution.

Based on the experimental analysis, we find that the remaining failures after applying S2C are not primarily caused by incorrect task-level semantic understanding or by the failure of the structured demonstration collection strategy itself. Instead, they mainly reflect the accumulation of execution-level errors. These errors are closely related to the hardware precision of the low-cost robotic platform, gripper stability, and unpredictable state variations in deformable-object manipulation.

\begin{center}
\captionof{table}{Failure event statistics in real-world robot evaluation under the S2C strategy.}
\label{tab:s2c_failure_statistics}
\begin{tabular}{lccc}
\toprule
Task & Grasp Failure & Wrong Action & Total Failure \\
\midrule
Block grasping and sorting & 50\% (1/2) & 50\% (1/2) & 2 \\
Towel folding & 90.47\% (19/21) & 9.53\% (2/21) & 21 \\
\bottomrule
\end{tabular}
\end{center}

\begin{figure}[H]
\centering
\setlength{\tabcolsep}{2pt}
\begin{tabular}{ccccc}

\multicolumn{5}{c}{\textbf{Grasp Instability}} \\
\includegraphics[width=0.18\linewidth]{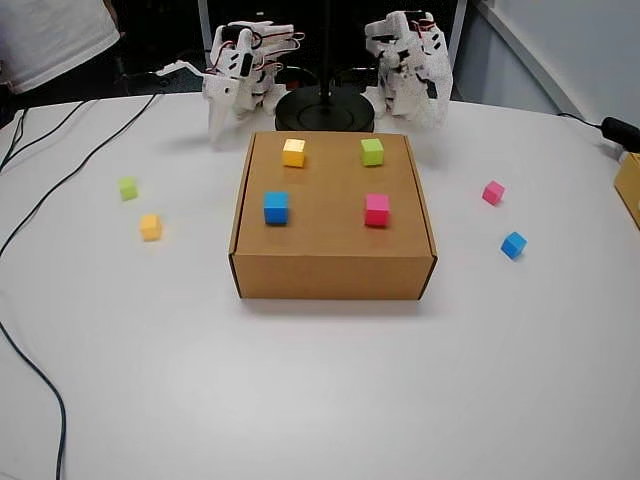}
&
\includegraphics[width=0.18\linewidth]{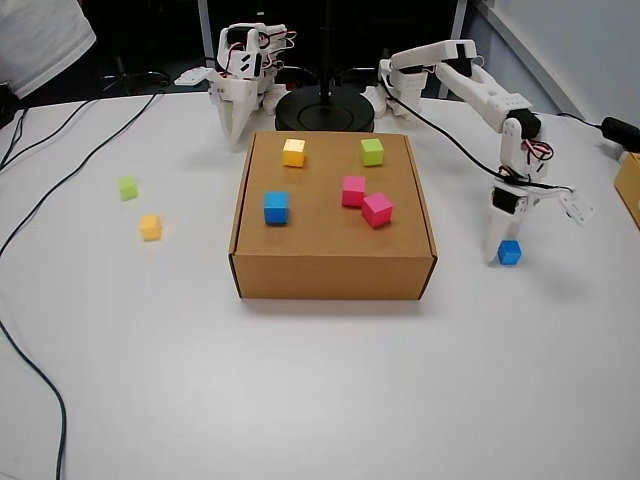}
&
\includegraphics[width=0.18\linewidth]{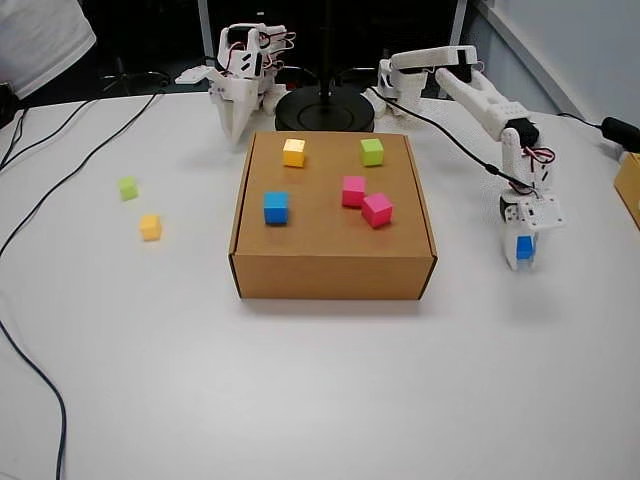}
&
\includegraphics[width=0.18\linewidth]{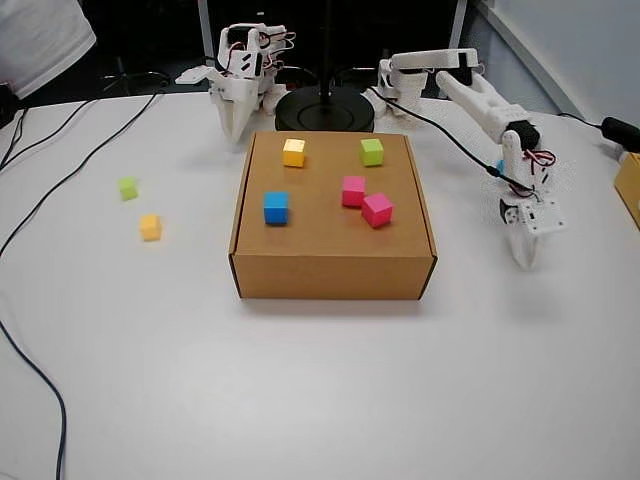}
&
\includegraphics[width=0.18\linewidth]{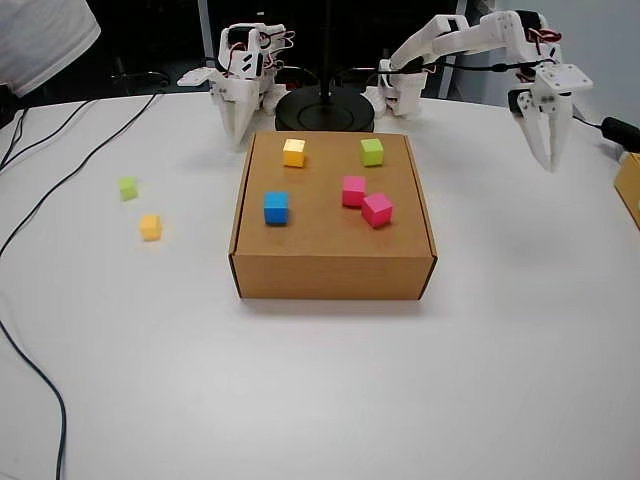}
\\[4pt]

\multicolumn{5}{c}{\textbf{Unstable Placement}} \\
\includegraphics[width=0.18\linewidth]{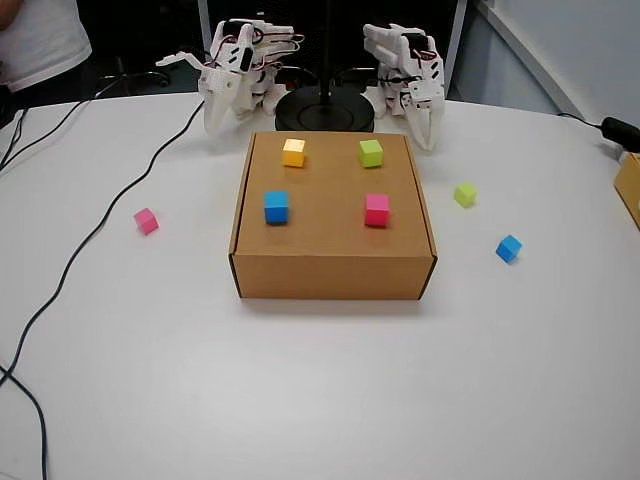}
&
\includegraphics[width=0.18\linewidth]{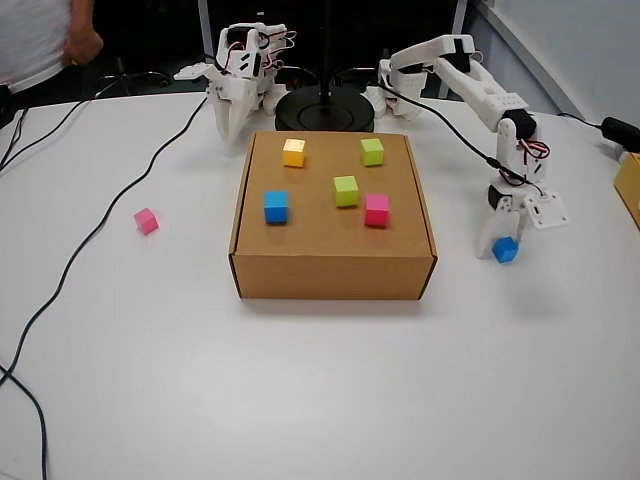}
&
\includegraphics[width=0.18\linewidth]{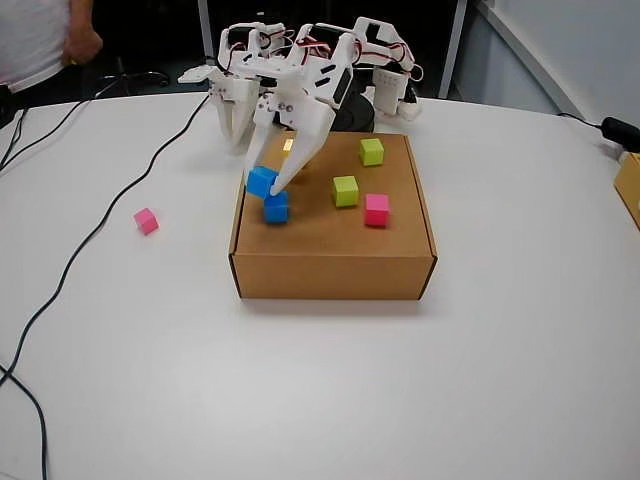}
&
\includegraphics[width=0.18\linewidth]{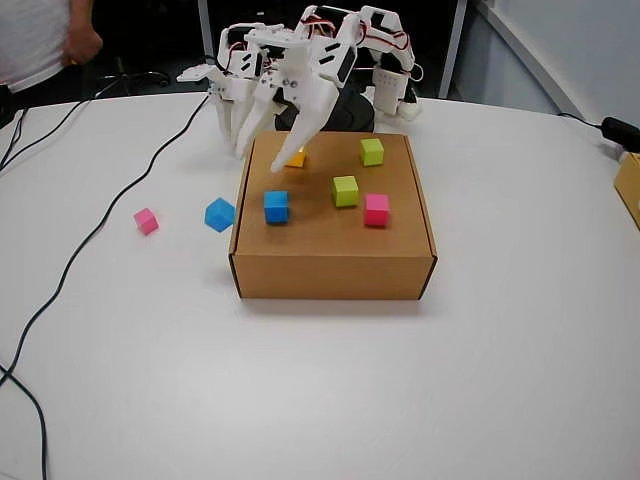}
&
\includegraphics[width=0.18\linewidth]{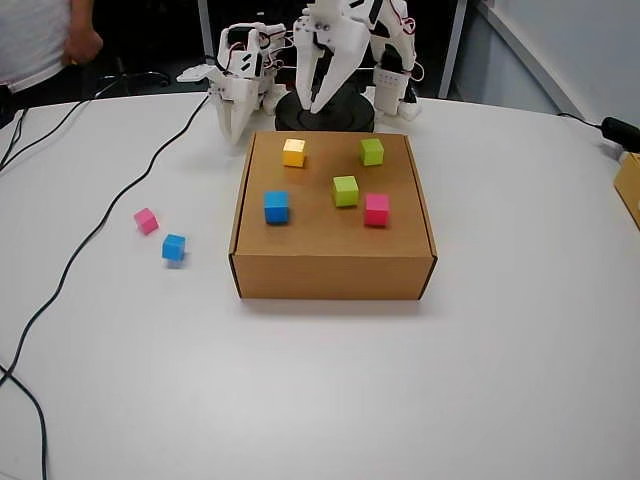}
\\

\end{tabular}
\caption{Representative failure cases in the block grasping and sorting task. Top: the robot successfully localizes the target block, but grasp instability causes the object to slip, be pushed away, or move out of the camera view. Bottom: the robot successfully grasps the block but fails to place it stably near the target region, resulting in placement failure.}
\label{fig:block_failure_cases}
\end{figure}

For the block grasping and sorting task, two main types of failures are observed during testing, as shown in Figure~\ref{fig:block_failure_cases}. The first type is grasp instability. In such cases, the manipulator is usually able to localize the target block and attempt the grasping action. However, due to unstable gripper closing positions, contact angles, or execution precision, the block may slip, be pushed away, or even move out of the camera view during grasping. This indicates that the policy has acquired a certain level of target localization and action initiation capability, but grasp execution can still fail due to the precision limitations of the low-cost robotic platform.

The second type is unstable or inaccurate placement. In such cases, the robot successfully grasps the target block but fails to place it stably near the target region, causing the block to fall into an unintended area or deviate from the expected position. This type of failure is usually not caused by a complete failure of color recognition, but rather by insufficient spatial precision and end-effector stability during placement. In dual-arm scenarios, such failures can sometimes be partially recoverable. For example, a placement error made by one manipulator may be partially compensated for by subsequent grasping or placement actions performed by the other manipulator.

\begin{figure}[H]
\centering
\setlength{\tabcolsep}{2pt}
\begin{tabular}{ccc}

\multicolumn{3}{c}{\textbf{Grasp Failure}} \\
\includegraphics[width=0.28\linewidth]{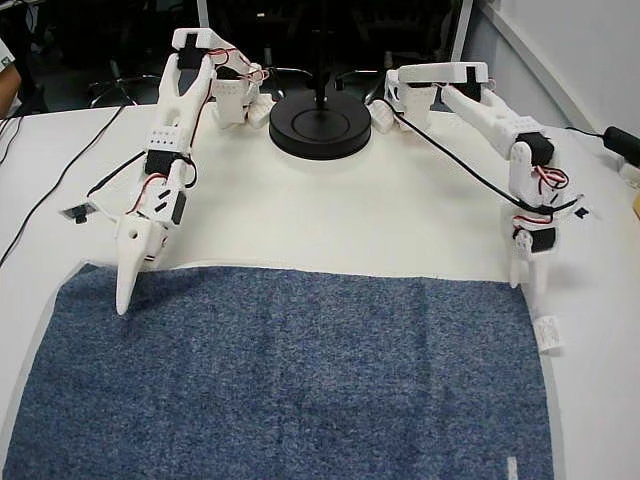}
&
\includegraphics[width=0.28\linewidth]{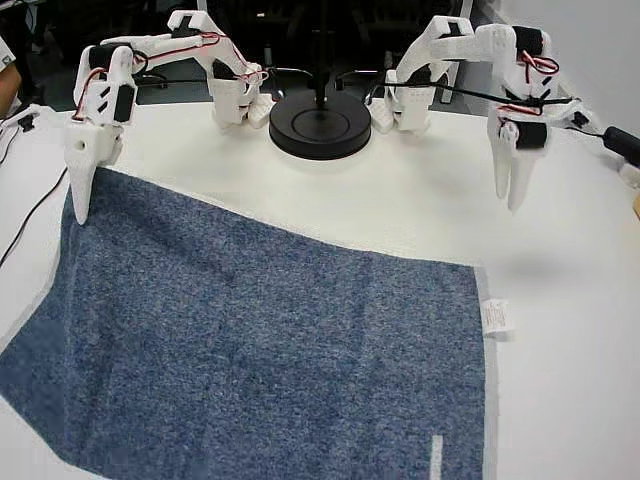}
&
\includegraphics[width=0.28\linewidth]{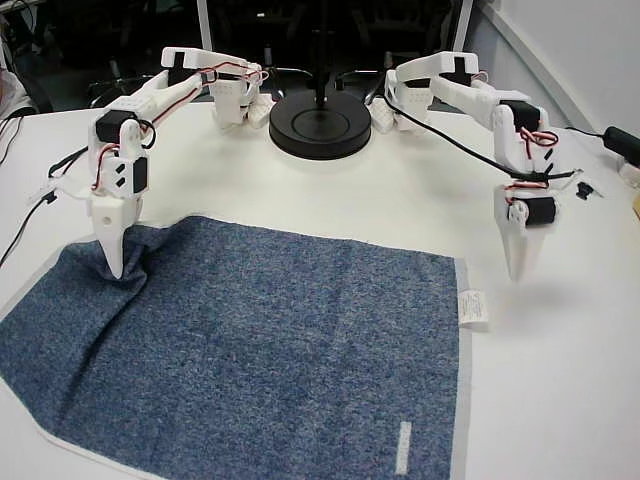}
\\[4pt]

\multicolumn{3}{c}{\textbf{Unpredictable Deformation}} \\
\includegraphics[width=0.28\linewidth]{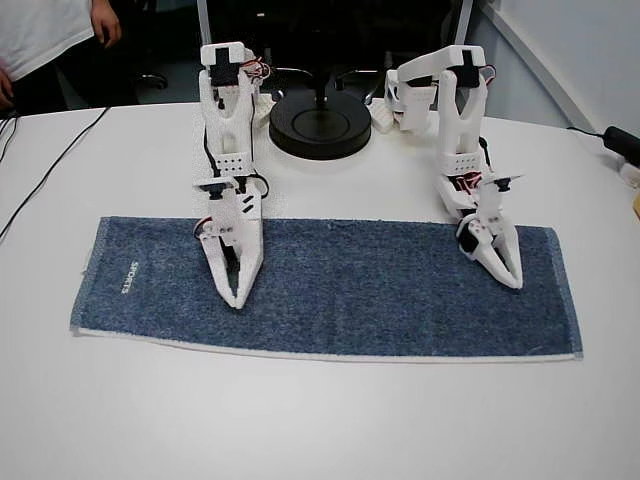}
&
\includegraphics[width=0.28\linewidth]{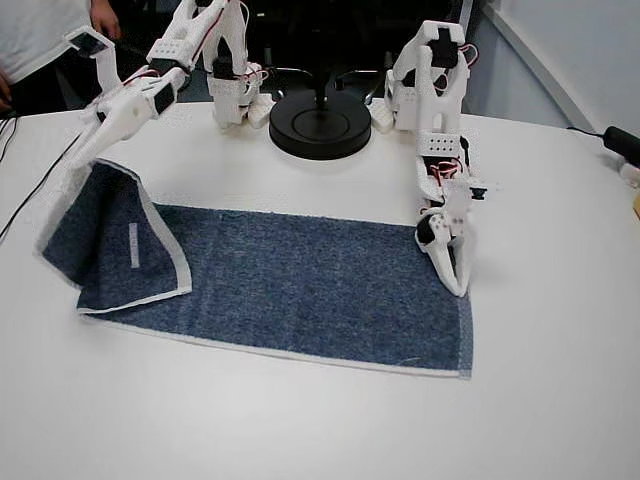}
&
\includegraphics[width=0.28\linewidth]{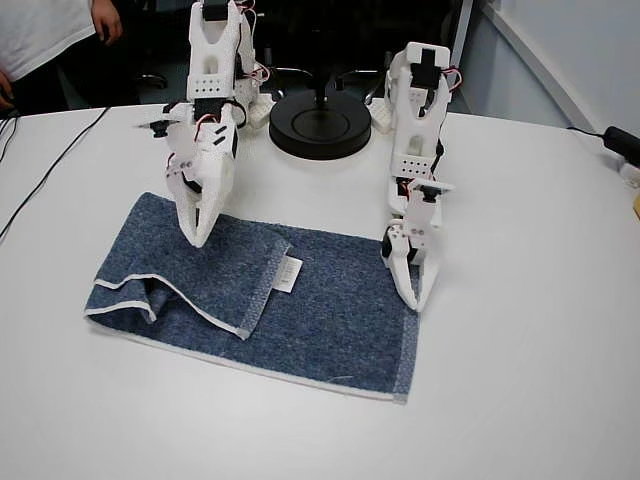}
\\

\end{tabular}
\caption{Representative failure cases in the towel-folding task. Top: the robot fails to stably grasp the towel, preventing subsequent state normalization and folding actions. Bottom: unpredictable deformation or slippage occurs during folding, causing the towel to enter intermediate states insufficiently covered by the training data and leading to hesitation, stopping, or incomplete folding.}
\label{fig:towel_failure_cases}
\end{figure}

For the towel-folding task, the failure types more directly reflect the difficulty of deformable-object manipulation, as shown in Figure~\ref{fig:towel_failure_cases}. The first type is initial grasp failure. Since the towel lies close to the tabletop and the SO-101 platform uses a two-finger gripper, it is difficult for the gripper to reliably insert under the towel edge and form a stable grasp. Once the initial grasp fails, the policy cannot proceed to the subsequent state normalization and folding stages, and the robot often remains stuck in the early stage of the task. This type of failure mainly reflects the physical limitation of low-cost hardware when grasping thin and soft objects.

The second type is unpredictable deformation during folding. Even when the robot successfully grasps and lifts the towel, subsequent folding actions may still produce intermediate states that are rarely covered in the training data due to deviations in grasp points, variations in gripping force, towel slippage, or local wrinkles. For deformable objects, small contact errors can be amplified into significant shape changes, making it difficult for the policy to infer the next action. This often appears as hesitation, stopping, or incomplete folding. These observations indicate that failures in towel folding arise not only from single-step execution errors, but also from error accumulation over long-horizon manipulation.

These failures indicate that, under low-cost robotic hardware and limited demonstration data, precise grasping, stable placement, and state variations in deformable objects remain major challenges. In particular, for long-horizon deformable-object manipulation tasks such as towel folding, future work may need to introduce more fine-grained intermediate stages, failure-recovery demonstrations, or additional data covering abnormal deformation states to improve robustness during complex execution.

\section{Discussion}

The experimental results demonstrate that the organization of demonstrations plays a critical role in policy learning under limited-data conditions. Rather than simply increasing the amount of collected data, the proposed S2C strategy improves learning efficiency by explicitly aligning the demonstration collection process with the dependency structure of the required manipulation capabilities. These findings suggest that, for long-horizon manipulation tasks, the structure and quality of demonstrations can be as important as their quantity.

A key characteristic of S2C is the progressive decomposition of complex manipulation tasks into capability-oriented learning stages. Instead of requiring the policy to simultaneously acquire low-level motor skills, perceptual reasoning, and high-level task sequencing from full demonstrations, S2C introduces these capabilities sequentially according to their dependency relationships. This staged learning process reduces optimization difficulty under limited-data conditions by allowing each stage to focus on a single capability while preserving previously acquired skills. The performance gains observed in the block grasping and sorting task further support the effectiveness of this capability-oriented demonstration collection strategy.

The towel-folding task further highlights both the effectiveness and limitations of the proposed strategy. Compared with direct end-to-end demonstration collection, S2C enables the policy to execute complete folding sequences in multiple trials, suggesting that the introduction of state normalization helps reduce the difficulty of learning long-horizon deformable-object manipulation. However, the overall success rate remains relatively low. This limitation is mainly associated with the intrinsic variability of deformable objects and the execution uncertainty of the low-cost robotic platform. Even after state normalization, visual and geometric variations may still arise during manipulation, and small execution errors can lead to intermediate states that are insufficiently covered by the available demonstrations.

The proposed framework can be instantiated for different object-centric long-horizon manipulation tasks using the general capability progression of basic manipulation, object perception and state understanding, and task execution. In this work, we evaluate S2C on block grasping and sorting, and towel folding, which represent rigid-object and deformable-object manipulation, respectively. The results suggest that this general decomposition rule is applicable to tasks with hierarchical task structures, multi-stage decision making, or strongly interdependent manipulation capabilities. 


\section{Future Work}

Although the proposed Simple-to-Complex (S2C) strategy improves policy learning under limited demonstration data, several limitations remain and motivate future research.

First, although S2C provides a general capability decomposition rule for object-centric long-horizon manipulation tasks, the concrete instantiation of each stage still needs to be specified for a given task. In this work, we instantiate the general progression of basic manipulation, object perception and state understanding, and task execution for block grasping and sorting and towel folding. Future work will investigate automatic stage instantiation and adaptive environment scheduling, enabling the robot to determine task-specific capability targets and scene complexity levels with less human intervention.

Second, the effectiveness of S2C may be related to the ability of VLA models to learn shared representations from structured and diverse demonstrations. Recent work on $\pi_{0.5}$ with egocentric human data suggests that sufficiently diverse pretraining can induce embodiment-agnostic representations and improve transfer across different embodiments and task settings~\cite{pi05_ego}. Inspired by this observation, our simple-to-complex demonstration strategy can be viewed as a way to expose the policy to progressively organized behavioral variations, encouraging the model to learn shared manipulation patterns across different stages. However, such shared representations are still learned implicitly through data-driven optimization. The model may capture statistical correlations between observations and actions without explicitly understanding the causal dependencies among actions, state transitions, and task outcomes.

Therefore, an important direction for future work is to integrate S2C with causal world modeling. Causal world models aim to move beyond purely predictive modeling by explicitly representing causal mechanisms, intervention effects, and counterfactual relations in embodied interaction~\cite{causal_world_model}. Incorporating such models into structured demonstration learning could allow the policy to reason not only about which action to execute, but also about why a particular action leads to a desired state transition. For long-horizon manipulation tasks, this may help the robot better understand the causal relationships between intermediate sub-skills, recover from execution errors, and generalize to unseen task configurations.

Third, deformable-object manipulation remains a major challenge under limited demonstration data. Although the proposed state normalization stage reduces the variability of towel configurations before folding, unpredictable deformation and execution errors can still lead to intermediate states that are insufficiently covered by the demonstrations. Future work will explore more fine-grained intermediate stages, failure recovery demonstrations, and causal reasoning mechanisms for deformable-object manipulation, so that the policy can better handle unexpected state changes during long-horizon execution.

Overall, future research will focus on extending S2C from a general structured demonstration strategy toward a more autonomous and causally grounded learning framework. By combining capability-oriented demonstration organization with causal world modeling, the robot may be able to acquire not only executable manipulation behaviors, but also a more explicit understanding of the causal structure underlying long-horizon task execution.

\bibliographystyle{unsrt}
\bibliography{sample-base}

\end{document}